\documentclass[lettersize,journal]{IEEEtran}
\usepackage{caption}
\usepackage{amsmath,amsfonts}
\usepackage{algorithm}
\usepackage{tabularray}
\usepackage{color}
\usepackage{array}
\usepackage[caption=false,font=normalsize,labelfont=sf,textfont=sf]{subfig}
\usepackage{textcomp}
\usepackage{stfloats}
\usepackage{float}
\usepackage{url}
\usepackage{verbatim}
\usepackage{graphicx}
\usepackage{cite}
\usepackage{subfig}
\usepackage[caption=false,font=normalsize,labelfont=sf,textfont=sf]{subfig}
\usepackage{caption}
\usepackage{subcaption}
\usepackage{makecell}
\usepackage{booktabs}
\usepackage{multirow}
\usepackage{algorithmicx}
\usepackage[noend]{algpseudocode}
\usepackage[table,xcdraw]{xcolor}
\usepackage[table,xcdraw]{xcolor}
\usepackage{color}
\usepackage{lineno}
\usepackage{nomencl}

\usepackage{hyperref} 

\usepackage{arydshln}

\usepackage{amsthm}   
\theoremstyle{plain}
\newtheorem{theorem}{Theorem}

\newtheorem{assumption}{Assumption}

\usepackage{cleveref}
\crefname{table}{Table}{Tables}
\crefname{assumption}{Assumption}{Assumptions}
\crefname{section}{Section}{Sections}
\crefname{appendix}{Appendix}{Appendixes}
\crefname{algorithm}{Algorithm}{Algorithms}
\crefname{figure}{Figure}{Figures}
\crefname{theorem}{Theorem}{Theorems}
\crefname{lemma}{Lemma}{Lemmas}
\crefname{definition}{Definition}{Definitions}
\renewcommand{\eqref}[1]{Eq.~(\textup{\ref{#1}})}

\usepackage{bm} 

\usepackage{algorithm}
\usepackage{algorithmicx}
\usepackage{algpseudocode}
\usepackage{amsthm}   
\theoremstyle{plain}

\usepackage{xcolor}
\usepackage{todonotes}

\usepackage{cleveref}
\crefname{table}{Table}{Tables}
\crefname{assumption}{Assumption}{Assumptions}
\crefname{section}{Section}{Sections}
\crefname{appendix}{Appendix}{Appendixes}
\crefname{algorithm}{Algorithm}{Algorithms}
\crefname{figure}{Figure}{Figures}
\crefname{theorem}{Theorem}{Theorems}
\crefname{lemma}{Lemma}{Lemmas}
\crefname{definition}{Definition}{Definitions}
\renewcommand{\eqref}[1]{Eq.~(\textup{\ref{#1}})}
\usepackage{bm} 

\makenomenclature

\begin{document}
\graphicspath{ {./picture/} }
\bibliographystyle{IEEEtran}

\title{Efficient Detection Framework Adaptation for Edge Computing: A Plug-and-play Neural Network Toolbox Enabling Edge Deployment }
\author{Jiaqi Wu, Shihao Zhang, Simin Chen, Lixu Wang, Zehua Wang, Wei Chen*, Fangyuan He, Zijian Tian, \textcolor{black}{F. Richard Yu,\textit{ Fellow IEEE}}, \textcolor{black}{Victor C. M. Leung, \textit{life Fellow IEEE}}
\thanks{This work
was supported by the National Natural Science Foundation of China under Grant 52074305, 52274160, 51874300 and the National Natural Science Foundation of China-Shanxi Provincial People's Government Coal-based Low Carbon Joint Fund U1510115, the Academic Research Projects of Beijing Union University (No.ZK90202106) and R$\&$D Program of Beijing Municipal Education Commission (KM202211417005). ({\emph{ Corresponding author: Wei Chen.}})
}

\thanks{Jiaqi Wu is with the School of Artificial Intelligence, China University of Mining and Technology (Beijing), Beijing 100083, China. The author is also with Department of Electrical and Computer Engineering, University of British Columbia, Vancouver, 2332 Main Mall Vancouver, BC Canada V6T 1Z4 (e-mail: wjq11346@student.ubc.ca)
}

\thanks{Simin Chen is with the university of Texas at dallas, 800 W Campbell Rd, Richardson, TX 75080. (e-mail:sxc180080@utdallas.edu)
}

\thanks{Lixu Wang is with Northwestern University at 633 Clark St, Evanston, IL 60208. (e-mail:lixuwang2025@u.northwestern.edu)
}

\thanks{Shihao Zhang, Zijian Tian are with the School of Artificial Intelligence, China University of Mining and Technology (Beijing), Beijing 100083, China.(e-mail: 286691859@qq.com; Tianzj0726@126.com)
}

\thanks{Wei Chen is with the School of Computer Science and Technology, China University of Mining and Technology, Xuzhou 221116, Jiangsu province, China. (e-mail: chenwdavior@163.com)
}

\thanks{Zehua Wang is with the Department of Electrical and Computer Engineering, The University of British Columbia, Vancouver, 2332 Main Mall Vancouver, BC Canada V6T 1Z4 (e-mail: zwang@ece.ubc.ca).}

\thanks{\textcolor{black}{{F. Richard Yu is with the Department of Systems and Computer Engineering,
Carleton University, Ottawa, ON K1S 5B6, Canada (e-mail: richard.yu@carleton.ca).}}}

\thanks{\textcolor{black}{{Victor C. M. Leung is with the Artificial Intelligence Research Institute, Shenzhen MSU-BIT University, Shenzhen 518172, China, the College of Computer Science and Software Engineering, Shenzhen University, Shenzhen 528060, China, and the  Department of Electrical and Computer Engineering, The University of British Columbia, Vancouver, BC V6T 1Z4,
Canada (e-mail: vleung@ieee.org).}}}

\thanks{Fangyuan He is with College of Applied Science and Technology, Beijing Union University, Beijing, 100012, China. (e-mail: yktfangyuan@buu.edu.cn)
}
}

\markboth{IEEE Transactions on Mobile Computing}%
{Shell \MakeLowercase{\textit{et al.}}: A Sample Article Using IEEEtran.cls for IEEE Journals}

\IEEEpubid{0000--0000/00\$00.00~\copyright~2021 IEEE}

\maketitle

\begin{abstract} Recently, edge computing has emerged as a prevailing paradigm in applying deep learning-based object detection models, offering a promising solution for time-sensitive tasks. However, existing edge object detection faces several challenges: 1) These methods struggle to balance detection precision and model lightweightness. 2) Existing generalized edge-deployment designs offer limited adaptability for object detection. 3) Current works lack real-world evaluation and validation. To address these challenges, we propose the \textit{\underline{E}dge \underline{D}etection \underline{Toolbox}} (ED-TOOLBOX), which leverages generalizable plug-and-play components to enable edge-site adaptation of object detection models. Specifically, we propose a lightweight \textit{Reparameterized Dynamic Convolutional Network} (Rep-DConvNet) that employs a weighted multi-shape convolutional branch structure to enhance detection performance. Furthermore, ED-TOOLBOX includes a \textit{Sparse Cross-Attention} (SC-A) network that adopts a localized-mapping-assisted self-attention mechanism to facilitate a well-crafted \textit{Joint Module} in adaptively transferring features for further performance improvement. In real-world implementation and evaluation, we incorporate an \textit{Efficient Head} into the popular \textit{You-Only-Look-Once} (YOLO) approach to achieve faster edge model optimization. Moreover, we identify that helmet detection---one of the most representative edge object detection tasks---overlooks band fastening, which introduces potential safety hazards. To address this, we build a \textit{Helmet Band Detection Dataset} (HBDD) and apply an edge object detection model optimized by the ED-TOOLBOX to tackle this real-world task. Extensive experiments validate the effectiveness of components in ED-TOOLBOX. In visual surveillance simulations, ED-TOOLBOX-assisted edge detection models outperform six \textit{state-of-the-art} methods, enabling real-time and accurate detection. These results demonstrate that our approach offers a superior solution for edge object detection.


\end{abstract}

\begin{IEEEkeywords}
Underground improper helmet-wearing detection, coal mine surveillance, Non-local attention network, Lightweight neural network, Edge computing
\end{IEEEkeywords}

\maketitle




\section{Introduction}

In recent years, the {practical application} of deep learning-based vision models has become a significant trend in the field of {artificial intelligence} \cite{wang2024yolov8, zhang2024generalist, chen2024transunet}.{ Object detection}, as a typical high-level visual analysis task, has extensive practical applications\cite{wu2024lightweight, jain2024fusion, wu2024small}, such as in security surveillance \cite{jain2024fusion} and power line inspection \cite{wu2024small}.

\begin{figure} [t!]
\centering
\includegraphics[width=0.51\textwidth,height=0.29\textwidth]{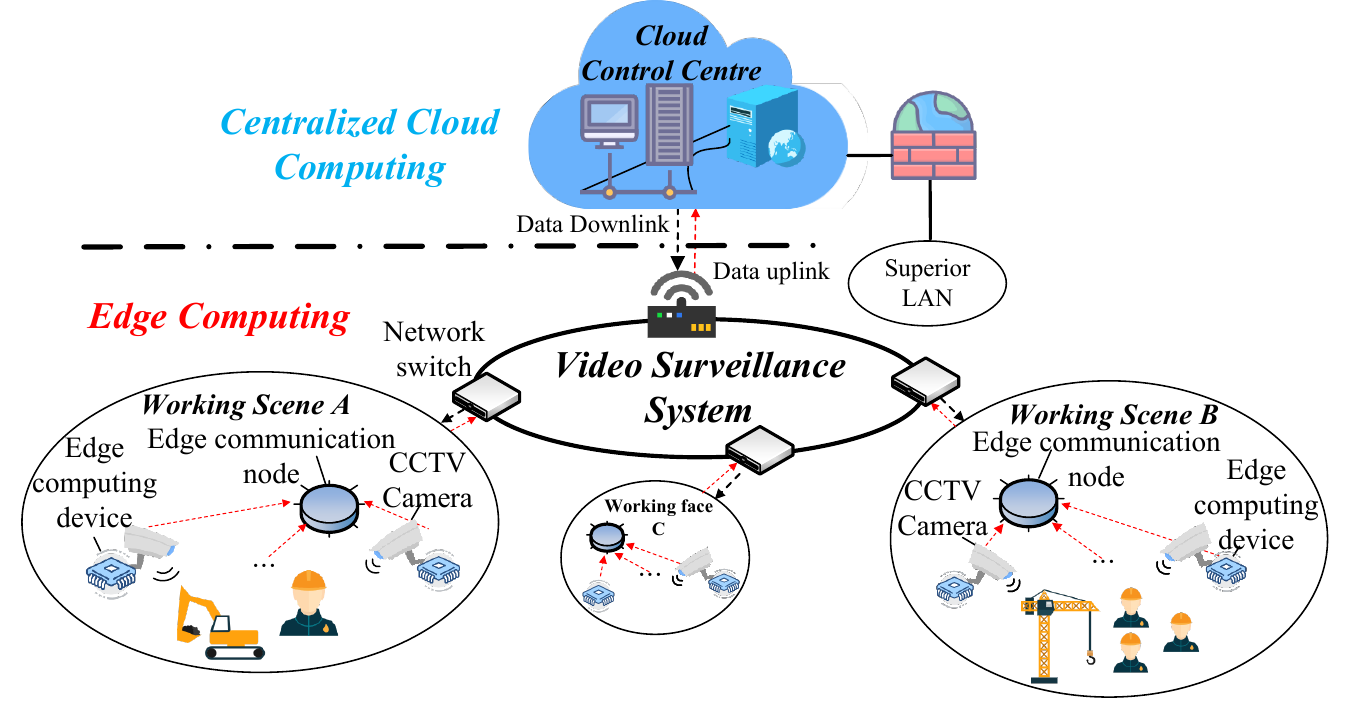}
\caption{The video surveillance system is based on an Internet of Video Things (IoVT) environment. The \textbf{black dashed line} distinguishes between \textcolor{blue}{Centralized Cloud Computing} and \textcolor{red}{Edge Computing}. In edge computing, data captured by \textit{CCTV Cameras} does not need to be uploaded to a \textit{Cloud Control Center}; instead, computation is performed directly on \textit{Edge Computing Devices} deployed in the \textit{Working Scenes}.}
\label{fig:edge&centralized}
\end{figure}

\textit{Edge computing} is an emerging computing paradigm \cite{hua2023edge, kong2022edge} that differs from traditional \textit{centralized cloud computing} \cite{bello2021cloud} by moving data analysis tasks closer to the data source, thereby reducing system latency caused by remote data transmission, as shown in Fig. \ref{fig:edge&centralized}. This characteristic makes it particularly well-suited as the deployment approach for detection models in time-sensitive visual tasks\cite{liang2022edge, wu2024small}. For example, this work \cite{wu2024lightweight} proposes a lightweight module, Rep-ShuffulNet, to assist You-Look-Only-Oncev8 (YOLOv8) in enabling edge deployment for a coal mine Internet of Video Things (IoVT) system. The work \cite{wu2024small} presents a distributed model deployment strategy for executing large deep learning models on edge devices with limited-resource within the IoVT system. Furthermore, the study \cite{yang2023cooperative} proposes an edge-empowered cooperative framework within the ECoMS system for real-time vehicle tracking and feature selection across multiple cameras.

However, existing edge detection methods still face the following \textit{issues}:

\begin{itemize} \item \textbf{Imbalance between performance and scale}: The performance of the original detection models significantly drops after lightweight edge deployment \cite{wu2024lightweight, zhao2024small}, making\\\\ \textit{it difficult to perform complex detection tasks}, especially for small object detection. 

\item \textbf{Lack of the detection-specifc designs}: Existing edge deployment solutions \cite{wu2024lightweight, howard2017mobilenets} \textit{lack detection-specific designs, limiting their ability to maintain detection performance} while enabling model deployment at the edge.

\item \textbf{Lack of real-world validation}: Existing researches \cite{mehta2022mobilevit, li2022efficientformer} primarily performs performance testing based on general or synthetic datasets, with a \textit{lack of evaluation on the effectiveness in real-world industrial tasks}.
\end{itemize}

To address these issues, we propose the \textit{Edge Detection Toolbox} (ED-TOOLBOX), which includes plug-and-play components to enable high-performance edge deployment of detection models. For modules in detection models, such as the Backbone, Neck, and Head modules commonly found in the YOLO framework \cite{ali2024yolo}, \textit{we propose specialized designs to preserve model performance}. Specifically, to maintain the feature extraction performance of edge detection models, we introduce a lightweight neural network, \textit{Reparameterized Dynamic Convolutional Network}
(Rep-DConvNet), to enhance the feature extraction network. We employ a multi-shape convolution branch structure with learnable weights to improve feature perception directionality and adaptability. Additionally, based on reparameterization strategies, it performs multi-channel fusion, \textit{significantly reducing the parameter count while maintaining feature extraction performance}. Additionally, we observe that existing detection frameworks focus on the construction of individual modules while overlooking the interconnections between them \cite{ali2024yolo, liu2016ssd}. To address this, we propose a novel spatial-channel attention structure \textit{with free-parameters to enable efficient key information transfer between modules}. This module is named the Joint Module, following the naming convention of the YOLO family \cite{ali2024yolo}. The \textit{Sparse Cross-Attention} (SC-A) module is a critical component, where we employ a multi-directional local computation strategy to \textit{reduce the complexity of capturing long-range dependencies while maintaining effectiveness}. Additionally, we design an Efficient Head to facilitate the decoupled prediction structure. It is worth noting that all components are plug-and-play, \textit{enabling compatibility with multiple detection frameworks}, such as the popular YOLO \cite{ali2024yolo} and SSD \cite{liu2016ssd} models, without significantly altering the network architecture. Furthermore, we are excited about the potential applications of the TOOLBOX to models for other tasks (as discussed in \textit{Section} \ref{sec:conlusion}).



\begin{figure} [b]
	\centering

            \subfloat{
		\includegraphics[width=3.5cm,height=2.5cm]{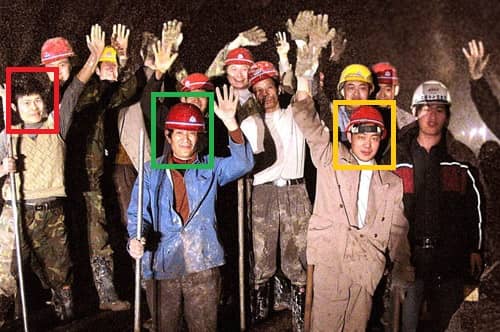}}
             \subfloat{
		\includegraphics[width=3.5cm,height=2.5cm]{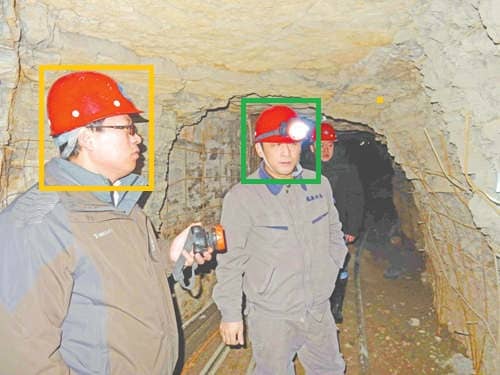}}
         \\
            \subfloat{
		\includegraphics[width=3.5cm,height=2.5cm]{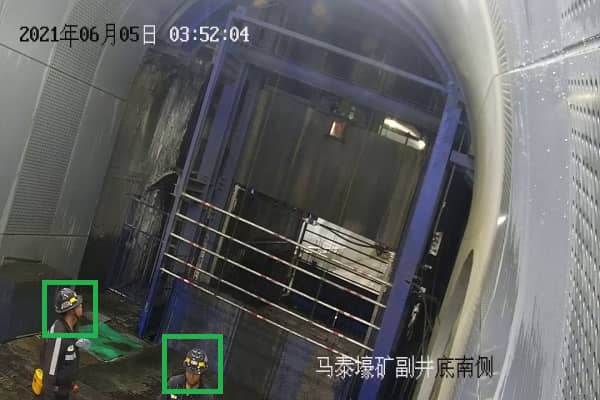}}
             \subfloat{
		\includegraphics[width=3.5cm,height=2.5cm]{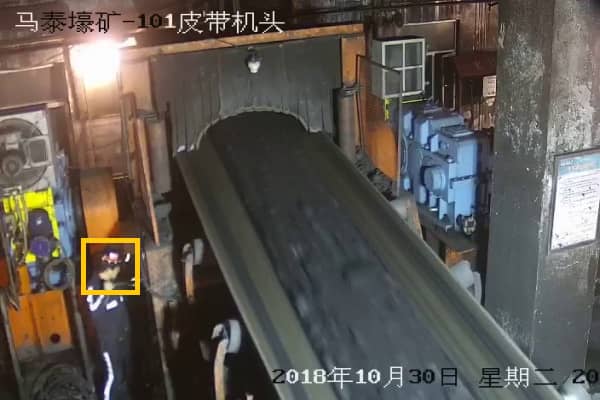}}

  \caption{Different helmet-wearing behaviours. 
Red boxes indicate \textit{no helmet worn}, green boxes indicate a \textit{helmet worn correctly}, and yellow boxes indicate a \textit{helmet worn but without the helmet band tied}. Not tying the hatband is a common improper helmet-wearing behaviour.}
\label{fig:strap}
\end{figure}

Furthermore, we find that existing helmet detection tasks \cite{Real-time, Deep-Learning} typically focus only on whether a helmet is worn, neglecting the helmet band, as shown in Fig. \ref{fig:strap}, which does not comply with safety regulations and introduces potential safety risks. To address this, we contribute a new \textit{Helmet Band Detection Dataset} (HBDD) and \textit{apply ED-TOOLBOX to implement edge detection for this real-world task}. In extensive experiments, we validate the effectiveness of various components in ED-TOOLBOX, demonstrating that they outperform existing state-of-the-art (SOTA) methods in feature perception and lightweight performance on both general datasets and our HBDD dataset. In visual surveillance simulations, ED-TOOLBOX enables edge deployment of detection models in the \textit{Internet of Video Things} (IoVT) system \cite{chen2020internet}, performing real-time and accurate small-sized strap detection. This highlights the practical application value of our approach. The contributions are summarized as follows:

\begin{itemize} 

\item \textbf{Overview}: We propose a \textit{Edge Detection Toolbox} (ED-TOOLBOX), which utilizes plug-and-play components to realize edge deployment of detection models with high performance.

\item \textbf{Key Components}: Our Rep-DConvNet module significantly reduces model size while maintaining feature extraction performance. The proposed SC-A module with free-parameters enables the adaptive transfer of crucial information between different modules in a detection model, with minimal computational cost.

\item \textbf{Practical Application Scenario}: We identify a neglected detection task in real-world industrial scenarios—helmet band detection. To address this, we contribute a \textit{Helmet Band Detection Dataset} (HBDD) and employ detection models assisted by ED-TOOLBOX.

\item \textbf{Evaluation}: Extensive experimental results demonstrate the effectiveness and practical application value of the components in ED-TOOLBOX.
\end{itemize}


\section{Related Work}
\label{Related Work}

\subsection{Edge Deployment Strategies}
\label{Edge Computing}

\textit{Edge Computing}\cite{hua2023edge, kong2022edge} is a distributed computing architecture that transfers computing power and data storage from the traditional cloud computing centre to a location closer to the data source. Compared with the \textit{Centralized Cloud Computing} \cite{bello2021cloud}, the edge computing approach can provide faster response, save network bandwidth, and also improve the reliability and stability of the system, as shown in Fig. \ref{fig:edge&centralized}. {Model Edge Deployment} is a critical issue, as it requires models to be adapted to resource-constrained embedded devices. Existing solutions include:







\textbf{Distributed Model Deployment Approaches.} These methods partition complex models and deploy them across multiple devices to enable edge computing. The study \cite{wu2024lightweight} proposes a model distribution deployment strategy for the edge deployment of complex algorithms in power IoVT systems, achieving real-time power line inspection. This work\cite{10018439} automates the partitioning of a convolutional neural network (CNN) model into submodels and generates the necessary code for their distributed execution to enable adaptive model partitioning. The study \cite{huang2023roofsplit} proposes a model distributed deployment approach based on roofline evaluation to maximize device resource utilization and improve model performance. \textit{However, these methods introduce additional communication overhead between devices, which significantly impacts the real-time performance of time-sensitive tasks, such as object detection.}


\textbf{Device-specific Designing Methods.} These methods design device-specific techniques based on the capabilities of the device to improve the device-model compatibility, with \textit{Neural Architecture Search} (NAS) \cite{ren2021comprehensive} being a typical example. HADAS\cite{bouzidi2023hadas} optimizes dynamic neural network architectures by jointly tuning the backbone and early exiting features on edge devices. Finch\cite{liu2023finch} accelerates federated learning by using hierarchical neural architecture search, dividing clients into clusters and sampling subnets in parallel. The study\cite{thomas2023neural} presents a neural architecture search algorithm to automatically optimize Transformer models for detecting various power system faults. \textit{However, these device-specific designing methods have drawbacks such as inefficiency and high costs.}

\begin{table*}[t]
\setlength{\tabcolsep}{0.08cm} 
\centering
\caption{Comparison of Edge Deployment Methods Based on Key Evaluation Points. The ``\checkmark" symbol indicates the presence of the corresponding feature, while the ``$\times$" symbol denotes its absence.}
\begin{tabular}{lcccccccc}
\toprule
\textbf{Method}     & \makecell{\textbf{Without} \\ \textbf{Communication} \\ \textbf{Overhead}} & \makecell{\textbf{Plug-and-Play} \\ \textbf{Capability}} & \makecell{\textbf{Structural} \\ \textbf{Optimization}} & \makecell{\textbf{Multiple} \\ \textbf{Optimization}} & \makecell{\textbf{Practical} \\ \textbf{Value}} & \makecell{\textbf{Adaptability}} & \makecell{\textbf{Task-Specific} \\ \textbf{ Optimization}} & \textbf{Task Type} \\ \midrule
\textbf{SIDD\cite{wu2024lightweight}}    & $\times$                           & $\times$                                  & \checkmark                 & $\times$                 & $\times$                           & \checkmark                       & $\times$                       & Power System Fault Detection           \\
\textbf{AS-CNN\cite{10018439}}      & $\times$                           & $\times$                                  & \checkmark                 & \checkmark                 & $\times$                           & $\times$                       & \checkmark                       & CNN-based Detection                \\
\textbf{HADAS\cite{bouzidi2023hadas}}      & \checkmark                           & \checkmark                        & \checkmark                           & \checkmark                 & $\times$                           & $\times$                      & \checkmark                       & Power System Fault Detection  \\
\textbf{Finch\cite{liu2023finch}}  & \checkmark                           & \checkmark                        & $\times$                           & \checkmark                 & $\times$                          & $\times$                       & \checkmark                       & Federated Learning             \\
\textbf{RoofSplit\cite{huang2023roofsplit}}      & $\times$                           & $\times$                                  & $\times$                           & $\times$                 & \checkmark                           & \checkmark                       & $\times$                       & Resource Utilization Optimization               \\
\textbf{NASformer\cite{thomas2023neural}}      & \checkmark                           & \checkmark                        & $\times$                           & \checkmark                 & \checkmark                           & $\times$                       & \checkmark                       & Transformer for Fault Detection                \\
\textbf{MobileNet\cite{howard2017mobilenets}}      & \checkmark                          & \checkmark                                  & \checkmark                 & $\times$                 & \checkmark                           & \checkmark                       & $\times$                       & Lightweight CNN                    \\
\textbf{ShuffleNet \cite{zhang2018shufflenet}}      & \checkmark                           & \checkmark                                  & \checkmark                 & $\times$                 & \checkmark                           & \checkmark                       & $\times$                       & Lightweight CNN            \\
\textbf{RepVGG \cite{ding2021repvgg}}      & \checkmark                           & \checkmark                                  & \checkmark                 & $\times$                 & \checkmark                           & \checkmark                       & $\times$                       & Lightweight CNN            \\
\textbf{Our method} & \checkmark                         & \checkmark                        & \checkmark                & \checkmark                 & \checkmark                & \checkmark                       & \checkmark                       & Edge-site detection              \\ \bottomrule
\end{tabular}
\label{tab:edgedeploymentcomparison}
\end{table*}


\textbf{Generalized Neural Networks}. These networks can replace regular CNNs to provide specific advantages, such as lightweight models. MobileNet\cite{howard2017mobilenets} is based on a streamlined architecture that utilizes depthwise separable convolutions to construct lightweight deep neural networks. ShuffleNet \cite{zhang2018shufflenet} is an efficient CNN architecture that reduces computational cost and improves accuracy by introducing two novel operations: group convolutions and channel shuffling. RepVGG \cite{ding2021repvgg} utilizes multiple channels during the model training phase to enhance feature extraction, and merges these channels during the inference phase to reduce the model size. \textit{However, these methods fail to effectively maintain the original performance, which limits their effectiveness for task-specific models.}


We propose a \textit{Edge Detection Toolbox} (ED-TOOLBOX) specifically designed for the edge deployment of detection models. It includes a variety of plug-and-play optimization components to adjust different modules of the detection models, such as the Backbone, Neck, and Head modules in YOLO framework \cite{ali2024yolo}. Compared to lightweight neural networks \cite{howard2017mobilenets, zhang2018shufflenet}, the detection-specific ED-TOOLBOX achieves better lightweighting while maintaining detection accuracy. In contrast to distributed model deployment approaches, this method does not introduce additional system latency. Unlike NAS methods \cite{bouzidi2023hadas, liu2023finch}, it offers adaptability for different devices and models. Additionally, we summarize the characteristics of the works mentioned above in TABLE \ref{tab:edgedeploymentcomparison} to further highlight the advantages of ED-TOOLBOX.

\subsection{Helmets Detection}
\label{Helmets Detection}

Helmets are crucial safety equipment, and proper helmet wearing effectively protects the lives of workers. Therefore, applying deep learning detection models to helmet detection tasks holds significant practical value. For example, the literature \cite{Real-time} first obtains the face boundary by regression algorithm and then detects the helmets in the bounding box. Similarly, literature \cite{Deep-Learning} first performs the motorbike boundary detection and then detects the helmets in this bounding box. Literature \cite{Pose-guided} first gain the head regions by designing feature points of the nose and ears and then detects the helmets in that cropped region. The literature \cite{Safety-helmet} uses the attention module to separate the foreground and background, and then realize fast and high-precision detection. The work \cite{Multi-Scale} continues the advantages of YOLOv3-Tiny’s small-volume, and uses hourglass residue modules to improve the detection accuracy. However, existing methods have several drawbacks: 

\begin{itemize} \item \textbf{Ignore Helmet-wearing Regulations}. They focus only on whether a helmet is worn, neglecting the necessity of securing the chin band, which not only creates safety hazards but also violates legal requirements. For example, the\textit{ Work Safety Law of China} (2021) \cite{wsl} stipulates that ``\textit{supervisory authorities may impose a fine when a helmet is worn without the chin band being secured.}" 

\item\textbf{Overlook Real-world Detection Needs}. In pursuit of high accuracy, complex network architectures are used, which reduce inference speed and hinder model deployment for real-time detection on edge devices. 

\item\textbf{Limited Detection Performance}. These models perform poorly on small-sized objects, making them unsuitable for helmet band detection.
\end{itemize}

To address these issues, we contribute a new \textit{Helmet Band Detection Dataset} (HBDD) and then conduct helmet band detection. Additionally, we propose a \textit{Edge detection Toolbox}, which includes a variety of plug-and-play components to optimize edge deployment for detection frameworks, such as YOLO family \cite{ali2024yolo} and SSD \cite{liu2016ssd}, enabling real-time small-sized helmet band detection.

\begin{figure*} [t]
	\centering
\includegraphics[width=0.9\textwidth,height=0.5\textwidth]{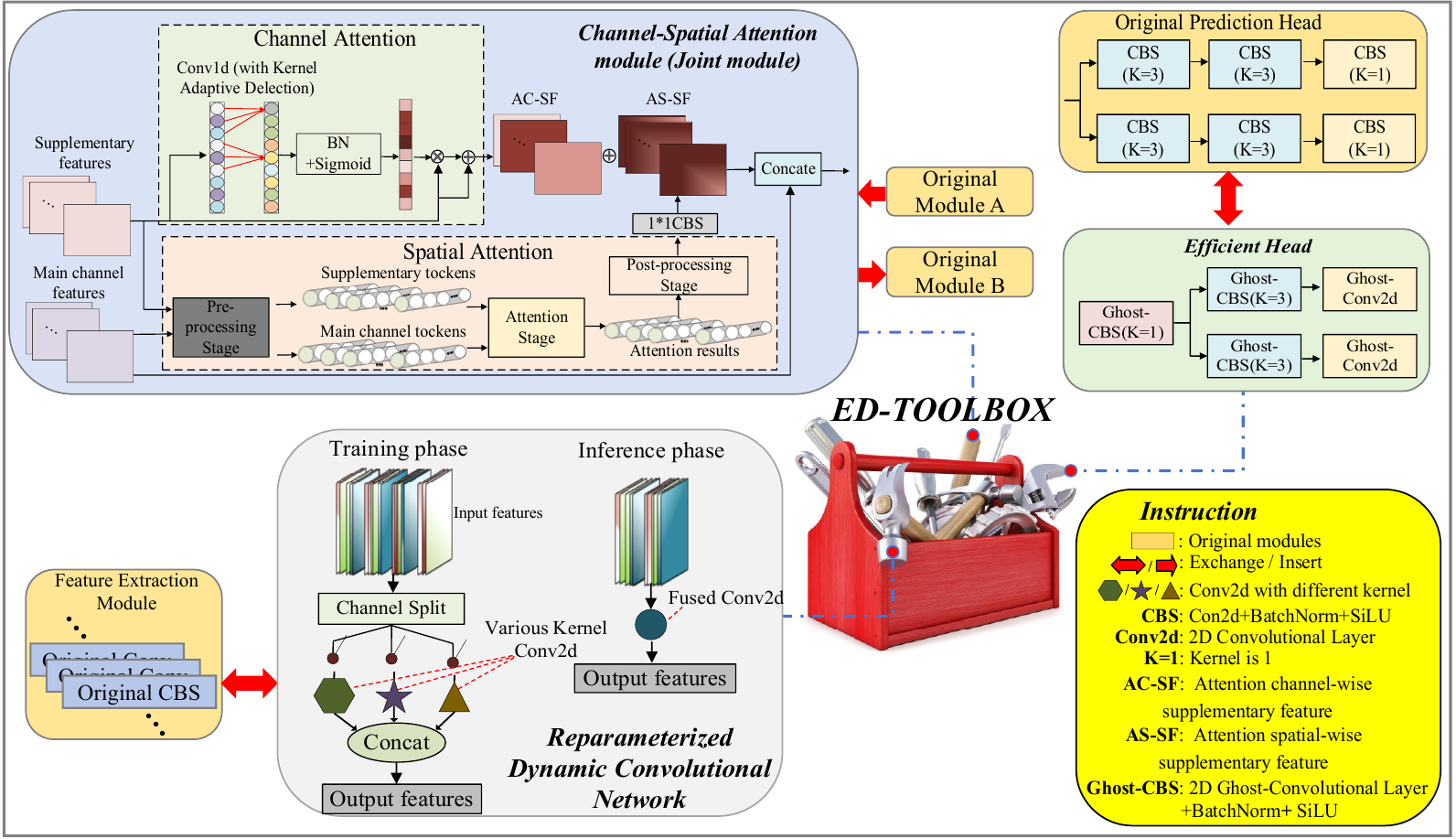}
  \caption{Overview of the ED-TOOLBOX. It includes the \textit{Reparameterized Dynamic Convolutional Network} (Rep-DConvNet), the \textit{Joint module}, and the \textit{Efficient Head}. These plug-and-play components achieve edge deployment of detection models and maintain excellent performance through ``replacement" and ``insertion."}
\label{fig:overview}
\end{figure*}

\section{Methodology}
\label{Methodology}

\subsection{Overview}
\label{Decentralized Coal Mine Video Surveillance System}


Inspired by real-world automotive upgrade kits, such as the BMW X5, which require no major modifications but can achieve city, off-road, and snow driving capabilities by swapping dedicated kits\footnote{\url{https://www.trailbuiltoffroad.com/vehicles/bmw/x5}}, we come up with the intuitive idea: \textit{Why not develop an ``edge detection kit" to assist the edge deployment of detection models?} Furthermore, we identify a limitation in general lightweight neural networks, which fail to effectively maintain the model’s perception performance \cite{howard2017mobilenets, zhang2018shufflenet}. Based on these insights, we design the \textit{Edge Detection Toolbox} (ED-TOOLBOX), specifically optimized for detection models, as shown in Fig. \ref{fig:overview}. This toolbox includes plug-and-play components, enabling the detection models to adapt to resource-constrained edge devices while maintaining excellent detection performance. The following sections will detail the key components.

\subsection{Reparameterized Dynamic Convolutional Network}
\label{section:rep}

\begin{figure} [t]
	\centering
\includegraphics[width=0.5\textwidth,height=0.3\textwidth]{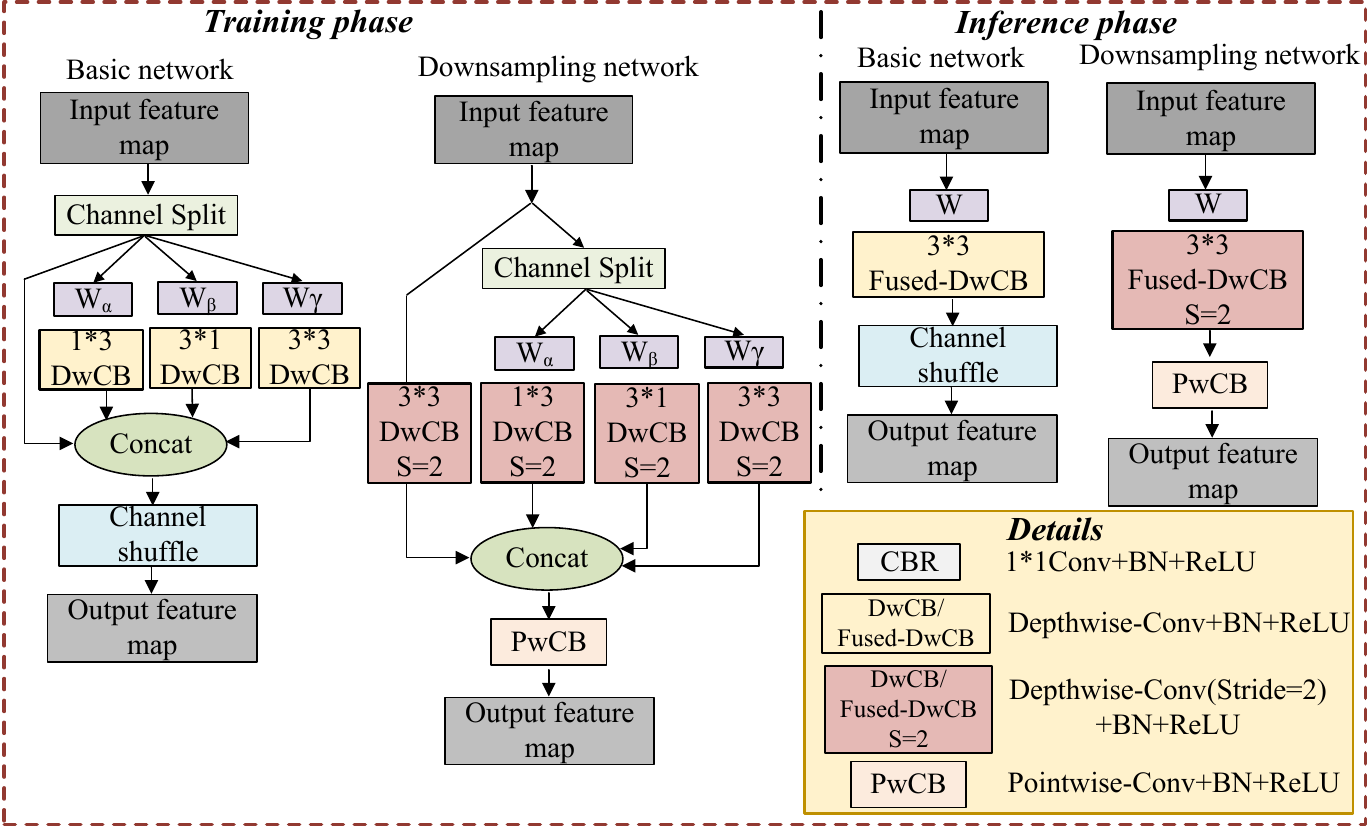}
  \caption{The structure of Rep-DConvNet. It decouples model \textit{Training phase} and \textit{Inference phase}, and is divided into a \textit{Basic network} and a \textit{Downsampling network} according to the design requirements of the detection framework \cite{ali2024yolo}.}
\label{fig:rep}
\end{figure}

Model lightweighting is a critical strategy for edge deployment \cite{wu2024lightweight, zhao2024small}.  Existing lightweight neural networks can reduce model size but often fail to maintain performance \cite{zhang2018shufflenet, howard2017mobilenets}. RepVGG \cite{ding2021repvgg} introduces a multi-channel architecture to preserve model performance; however, its simplified design hinders improvements in detection effectiveness. To address these issues, we propose a novel lightweight neural network, the \textit{Reparameterized Dynamic Convolutional Network\footnote{\url{https://github.com/word-ky/Edge-TOOLBOX/blob/main/backbone.py}}} ({Rep-DConvNet}), \textit{which focuses on maintaining detection performance}. Rep-DConvNet can be used to replace the standard convolutional layers in the feature extraction networks of detection models. According to the design requirements, Rep-DConvNet is divided into a \textit{Basic Network} for feature extraction and a \textit{Downsampling Network} for downsampling, as illustrated in Fig. \ref{fig:rep}. 

\noindent \textbf{Advantages.} (1) Unlike RepVGG \cite{ding2021repvgg}, Rep-DConvNet employs a weighted multi-shape convolutional branch structure to \textit{enhance feature perception directionality and adaptability, effectively improving detection performance}. (2) It utilizes depth-wise convolutional layers (DwConv) \cite{howard2017mobilenets}, channel splitting, and channel shuffle operations to replace conventional convolutional mappings, \textit{further reducing computational complexity}.

\noindent \textbf{Structure$\&$Processing.}  Taking the Basic Network as an example, the training process is outlined in Algorithm \ref{agrm:1}, and the details are explained as follows: Given an input feature map $\mathbf{X} \in \mathbb{R}^{C \times H \times W}$, where $C$, $H$, and $W$ denote the channels, height, and width, respectively, the feature map undergoes \textit{Channel-wise Division} into $n$ ($n$=3) groups along the channel dimension:
\begin{equation}
\mathbf{X} = \{\mathbf{X}_1, \mathbf{X}_2, \dots, \mathbf{X}_n\}, \quad \mathbf{X}_i \in \mathbb{R}^{\frac{C}{n} \times H \times W}, \quad \forall i \in [1, n].
\label{eq:divide}
\end{equation}

The $\mathbf{X}$ is processed through a \textit{Multi-channel Feature Extraction} mechanism. Feature extraction is conducted through multiple convolutional branches, including horizontal, vertical, square convolutions, and a residual connection. The outputs for each group are represented using the \(\text{\textbf{sgn}}\) function as follows:
\begin{equation}
\begin{aligned}
\mathbf{Y}^i = 
&\text{sgn}(i - 1) \cdot \sigma(\mathbf{W}^{1 \times 3} * \mathbf{X}_1) + 
\text{sgn}(i - 2) \cdot \sigma(\mathbf{W}^{3 \times 1} * \mathbf{X}_2) \\
&+ \text{sgn}(i - 3) \cdot \sigma(\mathbf{W}^{3 \times 3} * \mathbf{X}_3), \\ &\forall i \in \{\text{1: horizontal}, 
\text{2: vertical}, \text{3: square}\},
\label{eq:sgnconv}
\end{aligned}
\end{equation}
where $\mathbf{W}^{1 \times 3}$, $\mathbf{W}^{3 \times 1}$, and $\mathbf{W}^{3 \times 3}$ are the convolution kernels of DwConv for horizontal, vertical, and square convolutions, respectively. Here, $*$ denotes the convolution operation, and $\sigma(\cdot)$ is a non-linear activation function, such as ReLU. Additionally, the residual connection $\mathbf{X}$ directly forwards the input feature map to the output without alteration.

To adaptively emphasize significant features, inspired by conditional convolution strategies\cite{CondConv}, \textit{Weighting} is applied to the extracted features. Specifically, learnable weight matrices $\mathcal{W}_\alpha$, $\mathcal{W}_\beta$, and $\mathcal{W}_\gamma$ are assigned to the horizontal, vertical, and square convolutions, respectively. The weighted outputs are computed as $\mathbf{Z}^{\text{horizontal}} = \sigma(\mathcal{W}_\alpha *\mathbf{Y}^{\text{horizontal}})$, $\mathbf{Z}^{\text{vertical}} = \sigma(\mathcal{W}_\beta *\mathbf{Y}^{\text{vertical}})$, and $\mathbf{Z}^{\text{square}} = \sigma(\mathcal{W}_\gamma *\mathbf{Y}^{\text{square}})$.

Once all groups have been processed, the ``Concat" operation is performed, which involves concatenating the channels from the weighted convolutional branches and adding the residual channels element-wise:
\begin{equation}
\mathbf{Y}_{\text{concat}} = \text{Concat}(\mathbf{Z}_{\text{horizontal}}, \mathbf{Z}_{\text{vertical}}, \mathbf{Z}_{\text{square}}) + \mathbf{R},
\label{eq:concat}
\end{equation}
Where \( \mathbf{R} \) represents the residual channel, and \( \mathbf{Y}_{\text{concat}} \) is the output after channel concatenation, with \( \mathbf{Y}_{\text{concat}} \in \mathbb{R}^{C \times H \times W} \).

To enhance cross-channel interaction, a \textit{Channel Shuffle} operation is applied. The shuffle rearranges the channel order by dividing $\mathbf{Y}_{\text{concat}}$ into $G$ groups and interleaving the channels. This operation is mathematically expressed as:
\begin{equation}
\mathbf{Y}_{\text{output}}(c, h, w) = \mathbf{Y}_{\text{concat}}\left(g \cdot (c \bmod G) + \left\lfloor \frac{c}{G} \right\rfloor, h, w\right),
\label{eq:shuff}
\end{equation}
where $G$ is the number of channel groups, $g$ is the group index, and $c$ is the channel index. ``\textbf{mod}" is an abbreviation for ``\textbf{modulo}," representing the modulus operation. This \textit{Channel Shuffle} mechanism facilitates effective inter-group interaction and reduces computational complexity compared to conventional $1 \times 1$ convolutional mapping.

\begin{algorithm}[h]
\caption{Training process of Rep-DConvNet.}
\begin{algorithmic}[1]
\Require Input feature map $\mathbf{X} \in \mathbb{R}^{C \times H \times W}$, number of groups $n$, number of channel shuffle groups $G$.
\Ensure Output feature map $\mathbf{Y}_{\text{output}}$.

\State Divide $\mathbf{X}$ into $n$ groups along the channel dimension, in Eq. \ref{eq:divide} \Comment{\textit{Channel-wise Division}}

\For{each group $\mathbf{X}_i$}
    \State Perform horizontal, vertical, square convolutions, and residual connection, in Eq. \ref{eq:sgnconv} \Comment{\textit{Multi-channel Feature Extraction}}
    \State Perform \textit{Weighting} for each convolution branch:
    \[
\mathbf{Z}^{(k)} = \sigma(\mathcal{W}^{(k)} \mathbf{Y}^{(k)}), \quad \forall k \in \{\text{horizontal}, \text{vertical}, \text{square}\},
\] 

    \State Concatenate outputs from all groups, in Eq. \ref{eq:concat}
\EndFor
\State \textbf{end for}

\State Apply channel shuffle to rearrange the channel order using $G$ groups, in Eq. \ref{eq:shuff} \Comment{\textit{Channel Shuffle}}

\State \Return Output feature map $\mathbf{Y}_{\text{output}}$.
\end{algorithmic}
\label{agrm:1}
\end{algorithm}


\subsection{Sparse Cross-Attention Network}



\begin{figure} [t]
	\centering
\includegraphics[width=0.5\textwidth]{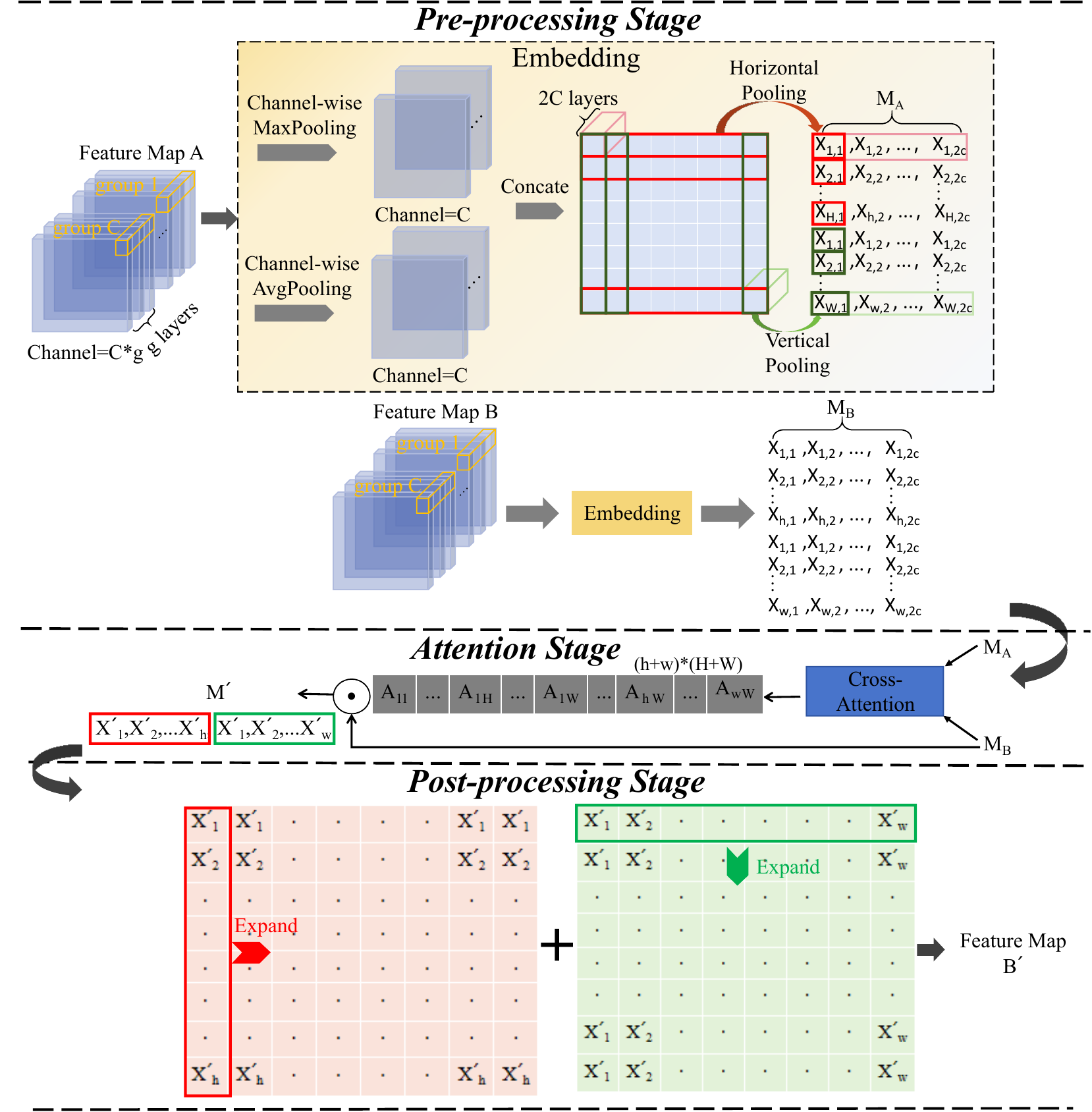}
  \caption{Procedure of Sparse Cross-Attention (SC-A). At the \textit{Pre-processing Stage}, it employs multiple pooling operations to perform local embedding mappings, reducing the complexity of subsequent computations. At the \textit{Attention Stage}, SC-A then performs self-attention computation on the two pooling results. Finally, at \textit{Post-processing Stage}, SC-A performs expansion operations to obtain the final attention result.}
\label{fig:SCA}
\end{figure}

We propose a novel \textit{Joint Module} to facilitate connections between original modules, named ``\textit{Joint}" following the YOLO family’s naming convention. The \textit{Joint Module} adopts a \textit{Channel-Spatial Attention\footnote{\url{https://github.com/word-ky/Edge-TOOLBOX/blob/main/neckv3.py}}} (CSA) structure \cite{guo2022attention}, where the channel-wise attention module utilizes ECANet \cite{wang2020eca}, and the spatial-wise attention module incorporates a \textit{Sparse Cross-Attention} (SC-A) mechanism based on the Non-local approach \cite{wang2018non}, as illustrated in Fig. \ref{fig:SCA}. 

\noindent \textbf{Advantages.} Compared to direct module connections, SC-A enables adaptive feature transfer with low computational complexity and no additional parameters: (1) \textit{Adaptive Connections}. Leveraging the Non-local mechanism \cite{wang2018non}, SC-A enhances downstream modules by focusing on multi-scale key features in the input feature maps. (2) \textit{Low Computational Complexity}. Building on previous studies \cite{Coordinate-attention, Strip-Pooling}, which demonstrated that Non-local-based structures with localized mapping can achieve outstanding performance without pointwise computations, we incorporate a more lightweight pooling operation into the Non-local network, further reducing computational complexity compared to convolutional operations. (3) \textit{Detection Task Adaptation}. Similar to the horizontal and vertical convolutions described in Section \ref{section:rep}, SC-A designs horizontal and vertical pooling to extract localized features, facilitating the detection of objects with diverse shapes in detection models.


\noindent \textbf{Structure$\&$Processing.} The detailed steps of SC-A are as follows, where Feature Map A / B represents the features of this module and the input features from the upstream module, respectively. Firstly, during the \textit{Pre-processing Stage}, the feature maps $\mathbf{A}$ and $\mathbf{B}$ are divided into $C$ groups along the channel dimension:
\begin{equation}
\mathbf{A} = \{\mathbf{A}_1, \mathbf{A}_2, \dots, \mathbf{A}_C\}, \quad \mathbf{B} = \{\mathbf{B}_1, \mathbf{B}_2, \dots, \mathbf{B}_C\},
\label{eq:divide2}
\end{equation}
where $\mathbf{A}_i, \mathbf{B}_i \in \mathbb{R}^{\frac{C}{n} \times H \times W}$ represent the groups of the feature maps. For each group, channel-wise max pooling and average pooling are performed to generate pooled feature maps: $\mathbf{P}_{\text{max}} = \text{MaxPool}(\mathbf{A}_i)$ and $\mathbf{P}_{\text{avg}} = \text{AvgPool}(\mathbf{A}_i)$, where $\mathbf{P}_{\text{max}}, \mathbf{P}_{\text{avg}} \in \mathbb{R}^{1 \times H \times W}$. These pooling results are concatenated along the channel dimension to form feature maps with $2C$ channels:

\begin{equation}
\mathbf{F} = \text{Concat}(\mathbf{P}_{\text{max}}, \mathbf{P}_{\text{avg}}), \quad \mathbf{F} \in \mathbb{R}^{2C \times H \times W}.
\label{eq:concat2}
\end{equation}

Subsequently, spatial-wise pooling is applied. For horizontal and vertical pooling, we compute: $\mathbf{T}_H = \text{Pool}_H(\mathbf{F})$, $\mathbf{T}_V = \text{Pool}_V(\mathbf{F})$, where $\mathbf{T}_H \in \mathbb{R}^{H \times 2C}$ and $\mathbf{T}_V \in \mathbb{R}^{W \times 2C}$. These outputs contain $H+W$ tokens (horizontal tokens and vertical tokens), with each token consisting of $2C$ elements. The concatenated tokens are denoted as $\mathbf{M}_A$ and $\mathbf{M}_B$, defined as:
\begin{equation}
\mathbf{M}_A = \{\mathbf{T}_H, \mathbf{T}_V\}, \quad \mathbf{M}_A \in \mathbb{R}^{(H+W) \times 2C},
\label{eq:MA}
\end{equation}
similarly, the input feature map $\mathbf{B}$ undergoes the same processing steps to obtain $\mathbf{M}_B \in \mathbb{R}^{(h+w) \times 2C}$. 

Secondly, \textit{Attention Stage} is performed between $\mathbf{M}_A$ and $\mathbf{M}_B$. The attention weights are calculated as:
\begin{equation}
\mathbf{A}_{\text{attn}}(i, j) = \text{Softmax}\left(\mathbf{M}_A(i) \cdot \mathbf{M}_B(j)^\top\right),
\label{eq:attention}
\end{equation}
where $\mathbf{A}_{\text{attn}} \in \mathbb{R}^{(h+w) \times (H+W)}$ represents the attention values. These weights are used to compute the attention-enhanced tokens $\mathbf{M}' = \mathbf{A}_{\text{attn}}(j) \cdot \mathbf{M}_A$, where $\mathbf{M}' \in \mathbb{R}^{2C}$ contains the dependencies between token $j$ in $\mathbf{M}_B$ and all tokens in $\mathbf{M}_A$.

\begin{algorithm}[h!]
\caption{SC-A Processing Steps}
\label{alg:SC-A}
\begin{algorithmic}[1]
\Require Input feature maps $\mathbf{A} \in \mathbb{R}^{C \times H \times W}$ and $\mathbf{B} \in \mathbb{R}^{C \times h \times w}$, number of groups $C$, spatial dimensions $H, W, h, w$.
\Ensure Output feature map $\mathbf{B}' \in \mathbb{R}^{C \times h \times w}$.

\State \textit{Pre-processing Stage}
\State Divide $\mathbf{A}$ and $\mathbf{B}$ into $C$ groups in Eq. \ref{eq:divide2},

\For{each group $\mathbf{X}_i \in \{\mathbf{A}_i, \mathbf{B}_i\}$}
    \State Perform max pooling and average pooling:
    \[
    \mathbf{P}_{\text{max}} = \text{MaxPool}(\mathbf{X}_i), \quad \mathbf{P}_{\text{avg}} = \text{AvgPool}(\mathbf{X}_i).
    \]
    \State Concatenate pooling results in Eq. \ref{eq:concat2}
    \State Perform horizontal and vertical pooling:
    \[
    \mathbf{T}_H = \text{Pool}_H(\mathbf{F}), \quad \mathbf{T}_V = \text{Pool}_V(\mathbf{F}).
    \]
    \State Concatenate tokens $\mathbf{M}_A$ in Eq. \ref{eq:MA}, \Comment{Same as for $\mathbf{M}_B$.}
\EndFor

\State \textit{Attention Stage}
\State Compute attention weights between $\mathbf{M}_A$ and $\mathbf{M}_B$, as shown in Eq. \ref{eq:attention},
\State Compute attention-enhanced tokens:
\[
\mathbf{M}' = \mathbf{A}_{\text{attn}}(j) \cdot \mathbf{M}_A.
\]

\State \textit{Post-processing Stage}
\State Expand tokens $\mathbf{M}'$ into original spatial dimensions:
\[
\mathbf{F}'_H(h, w) = \mathbf{M}'_h, \quad \mathbf{F}'_V(h, w) = \mathbf{M}'_v.
\]
\State Perform pixel-wise addition to obtain the final output B' in Eq. \ref{eq:addition}.

\State \Return Final output feature map $\mathbf{B}'$.
\end{algorithmic}
\end{algorithm}

Finally, in \textit{Post-processing Stage}, the attention-enhanced tokens $\mathbf{M}'$ are expanded back into the original spatial dimensions. For horizontal and vertical expansions, we compute $\mathbf{F}'_H(h, w) = \mathbf{M}'_h, \quad \mathbf{F}'_V(h, w) = \mathbf{M}'_v$, where $\mathbf{F}'_H \in \mathbb{R}^{H \times W \times 2C}$ and $\mathbf{F}'_V \in \mathbb{R}^{H \times W \times 2C}$ are the expanded feature maps. These are combined with pixel-wise addition to obtain the final output:
\begin{equation}
\mathbf{B}' = \mathbf{F}'_H + \mathbf{F}'_V,
\label{eq:addition}
\end{equation}
where each pixel in $\mathbf{B}'$ captures dependencies with all pixels in $\mathbf{A}$. In other words, the importance of upstream input features $\mathbf{B}'$ is determined based on the requirements of the current module features $\mathbf{A}$.

\subsection{Efficient Head}



The YOLO series \cite{ali2024yolo} is one of the most popular object detection algorithms, widely applied to real-world detection tasks \cite{wu2024lightweight, zhao2024small}. In its latest versions (e.g. v8), the decoupled \textit{Prediction Head} has demonstrated outstanding performance. However, its dual-branch structure significantly increases the model size. Inspired by \cite{YOLOv6}, we propose an \textit{Efficient Head}.

\noindent \textbf{Advantages.} (1) \textit{Lightweight Structure}. Compared to the \textit{Prediction Head} in YOLOv8, we introduce a single-branch structure to perform holistic feature mapping, as shown in Fig. \ref{fig:overview}. To further reduce the parameter count, we replace the original CBS modules in the head with Ghost-CBS modules \cite{han2020ghostnet}. (2) \textit{Edge Detection Adaptation}. To address the limitations of edge devices, particularly for small-object detection, we add a new detection head to the original structure, processing feature maps of size 160×160. This significantly improves the detection performance for small-scale objects.

\section{Analysis on Computing Complexity}
\label{sec:Computing Complexity}


\textit{Computational complexity} is a crucial metric for assessing the feasibility of deploying detection models on resource-constrained embedded devices \cite{howard2017mobilenets}. Beyond empirical experiments in \textit{Section} \ref{sec: experiment}, we perform theoretical derivations to analyze the computational complexity of key components in ED-TOOLBOX. The detailed analysis is as follows:


\begin{assumption} 
\label{assu1:condition}
The multi-branch network consists of four branches: one residual connection and three convolutional branches, each employing a \(3 \times 3\) convolutional layer.
\end{assumption}

\begin{assumption} 
\label{assu2:condition}
The number of input and output channels (\(C_{\text{in}}\) and \(C_{\text{out}}\)) are assumed to be equal, following the structure of the Basic Network in the YOLO framework, as illustrated in Fig. \ref{fig:YOLODBtoolbox}.

\end{assumption}

\begin{assumption} 
\label{assu3:condition}
In the Channel Shuffle, the parameter \(c\) is a small positive value, according to the low computational complexity characteristic of indexing, reordering, and any associated memory operations \cite{zhang2018shufflenet}.
\end{assumption}

\begin{theorem} 
\label{theo:1} 
Under \cref{assu1:condition,assu2:condition,assu3:condition}, the computational complexity of Rep-DConvNet is lower than that of the original RepVGG and regular convolutional layers.
\end{theorem}

\textit{Proof.}

\textbf{The Training Phase.} The total computational complexity of {RepVGG} denoted as \( \mathcal{C}_{\text{train}} \), comprises the complexities of the three convolutional branches (\( \mathcal{C}_{\text{conv}} \)) and the residual connection (\( \mathcal{C}_{\text{residual}} \)) as:
\begin{equation}
\begin{aligned}
\mathcal{C}_{\text{train}} &= 3*\mathcal{C}_{\text{conv}} + \mathcal{C}_{\text{residual}}\\
&=3 * (9 * C_{\text{in}} * C_{\text{out}} * H * W) + C_{\text{out}} * H * W\\
&=C_{\text{out}} * H * W(27*C_{\text{in}}+1),
\end{aligned}
\end{equation}
where, \( C_{\text{in}} \) and \( C_{\text{out}} \) represent the number of input and output channels, respectively; \( H \) and \( W \) denote the height and width of the feature map; \( 9 \) corresponds to the number of operations in each \(3 \times 3\) convolutional kernel.

The total computational complexity \(C_{\text{train}}\) of our {Rep-DConvNet} includes contributions from the convolution branches \(C_{\text{branches}}\), weighting \(C_{\text{weighting}}\), the residual connection \(C_{\text{residual}}\), and the channel shuffle \(C_{\text{shuffle}}\) operation. And the parameter \(c\) represents the computational cost per channel.
\begin{equation}
\begin{aligned}
C_{\text{train}} &= C_{\text{branches}} + C_{\text{weighting}} + C_{\text{residual}} + C_{\text{shuffle}}\\
&+ C_{\text{out}} \cdot H \cdot W + c \cdot C_{\text{out}}\\
&= 9 \cdot C_{\text{in}} \cdot H \cdot W + 4 \cdot C_{\text{out}} \cdot H \cdot W +c \cdot C_{\text{out}}
\end{aligned}
\end{equation}

The \textbf{computational complexity ratio} between {Rep-DConvNet} and {RepVGG} is provided below. \textit{As \(C_{\text{in}}\) increases, the value of \(\lambda_{\text{train}}\) decreases, indicating that \textit{Rep-DConvNet} exhibits lower complexity, which is beneficial for improving training efficiency.}
\begin{equation}
\begin{aligned}
\lambda_{\text{train}} &= \frac{C_{\text{Rep-DConvNet}}}{C_{\text{RepVGG}}}\\
&= \frac{9 \cdot C_{\text{in}} \cdot H \cdot W + 4 \cdot C_{\text{out}} \cdot H \cdot W + c \cdot C_{\text{out}}}{C_{\text{out}} \cdot H \cdot W \cdot (27 \cdot C_{\text{in}} + 1)}\\
&= \frac{9 \cdot C_{\text{in}} + 4 \cdot C_{\text{out}} + \frac{c \cdot C_{\text{out}}}{H \cdot W}}{C_{\text{out}} \cdot (27 \cdot C_{\text{in}} + 1)}\\
&\overset{\mathbf{A}}{=} \frac{9 \cdot C_{\text{in}} + 4 \cdot C_{\text{out}}}{C_{\text{out}} \cdot (27 \cdot C_{\text{in}} + 1)}\\
&\overset{\mathbf{B}}{=} \frac{9 + 4}{27 \cdot C_{\text{in}} + 1},
\label{eq:lamdatrain1}
\end{aligned}
\end{equation}
where, Strategy \(\mathbf{A}\) denotes that the term \(\frac{c \cdot C_{\text{out}}}{H \cdot W}\) is negligible, following {Assumption} \ref{assu3:condition}. Strategy \(\mathbf{B}\) follows {Assumption} \ref{assu2:condition}.

\textbf{The Inference Phase.} {RepVGG} fuses all branches into a single equivalent \(3 \times 3\) convolutional layer, the total computational complexity denotes:
\begin{equation}
\mathcal{C}_{\text{infer}} = 9 \cdot C_{\text{in}} \cdot C_{\text{out}} \cdot H \cdot W,
\end{equation}
the computational complexity of a \textit{regular convolutional layer} is equivalent to that of the fused layer.

The total computational complexity of our {Rep-DConvNet} during the inferece phase:
\begin{equation}
\begin{aligned}
C_{\text{infer}} &= C_{\text{branches}} +  C_{\text{weighting}} + C_{\text{shuffle}}\\
&= 9 \cdot C_{\text{in}} \cdot H \cdot W + 3 \cdot C_{\text{out}} \cdot H \cdot W +c \cdot C_{\text{out}}
\end{aligned}
\end{equation}

The \textbf{computational complexity ratio} is given as follows, demonstrating that \textit{Rep-DConvNet has significantly lower computational complexity compared to {RepVGG}. Moreover, as \(C_{\text{in}}\) increases, the gap between the two becomes even larger.}
\begin{equation}
\begin{aligned}
\lambda_{\text{infer}} &= \frac{\mathcal{C}_{\text{Rep-DConvNet}}}{\mathcal{C}_{\text{RepVGG}}}\\
&=\frac{9 \cdot C_{\text{in}} \cdot H \cdot W + 3 \cdot C_{\text{out}} \cdot H \cdot W + c \cdot C_{\text{out}}}{9 \cdot C_{\text{in}} \cdot C_{\text{out}} \cdot H \cdot W}\\
&=\frac{9 \cdot C_{\text{in}} + 3 \cdot C_{\text{out}} + \frac{c \cdot C_{\text{out}}}{H \cdot W}}{9 \cdot C_{\text{in}} \cdot C_{\text{out}}}\\
&\overset{\mathbf{A}}{=}\frac{9 \cdot C_{\text{in}} + 3 \cdot C_{\text{out}}}{9 \cdot C_{\text{in}} \cdot C_{\text{out}}}\\
&\overset{\mathbf{B}}{=}\frac{4}{3 \cdot C_{\text{in}}}\
\end{aligned}
\end{equation}
where, Strategy \(\mathbf{A}\) aligns with {Assumption} \ref{assu3:condition}, and Strategy \(\mathbf{B}\) aligns with {Assumption} \ref{assu2:condition}, consistent with Eq. \ref{eq:lamdatrain1}.

\begin{assumption} 
\label{assu4:condition}
According to the detection model structure (see Fig. \ref{fig:YOLODBtoolbox}), the input feature maps \( \mathbf{A} \) and \( \mathbf{B} \) have identical dimensions, i.e., \( H = h \) and \( W = w \).  
\end{assumption}

\begin{assumption} 
\label{assu5:condition}
Based on practical experience in detection tasks \cite{ali2024yolo}, the aspect ratio of input images is typically close to 1. For simplicity, we assume aspect ratio of 1 (see Fig. \ref{fig:YOLODBtoolbox}).
\end{assumption}

\begin{theorem} 
\label{theo:2} 
Under \cref{assu4:condition,assu5:condition}, the computational complexity of SC-A is lower than that of the original Non-Local network.
\end{theorem}

\textit{Proof.}

The computational complexity of \textbf{SC-A} is derived in three stages: \textit{Pre-processing Stage}, \textit{Attention Stage}, and \textit{Post-processing Stage}, assuming feature maps \( \mathbf{A} \in \mathbb{R}^{C \times H \times W} \) and \( \mathbf{B} \in \mathbb{R}^{C \times h \times w} \).

At the \textit{Pre-processing Stage}, the input feature maps \( \mathbf{A} \) and \( \mathbf{B} \) are divided into \( C \) groups along the channel dimension, and the cost of this division is negligible. For each group, max pooling and average pooling operations are performed. The complexity for these operations is \( \mathcal{C}_{\text{pool}} = 2 \cdot C \cdot H \cdot W + 2 \cdot C \cdot h \cdot w \). Subsequently, horizontal pooling (\( \text{Pool}_H \)) and vertical pooling (\( \text{Pool}_V \)) are applied, with a complexity of \( \mathcal{C}_{\text{spatial}} = 2 \cdot C \cdot H \cdot W + 2 \cdot C \cdot h \cdot w \). Combining these, the total pre-processing complexity is:  
\begin{equation}
\mathcal{C}_{\text{pre\_stage}} = 4 \cdot C \cdot H \cdot W + 4 \cdot C \cdot h \cdot w.
\end{equation}

At the \textit{Attention Stage}, attention weights are computed between tokens from the feature maps. The complexity of computing attention weights is \( \mathcal{C}_{\text{attn}} = (H + W) \cdot (h + w) \cdot (2C) \). Multiplying the attention weights with tokens adds another complexity term of \( \mathcal{C}_{\text{mult}} = (H + W) \cdot (h + w) \cdot (2C) \). Therefore, the total complexity for the attention stage is: 
\begin{equation}
\mathcal{C}_{\text{attn\_stage}} = 2 \cdot (H + W) \cdot (h + w) \cdot (2C).
\end{equation}

At the \textit{post-processing stage}, horizontal and vertical tokens are expanded back into the spatial dimensions. The expansion complexity is \( \mathcal{C}_{\text{expand}} = 2 \cdot H \cdot W \cdot 2C \). Finally, an element-wise addition is performed, with a complexity of \( \mathcal{C}_{\text{add}} = H \cdot W \cdot 2C \). Thus, the total post-processing complexity is:  
\begin{equation}
\mathcal{C}_{\text{post\_stage}} = 6 \cdot H \cdot W \cdot C.
\end{equation}

The \textit{total computational complexity} of SC-A is the sum of all three stages:
\begin{equation}
\begin{aligned}
\mathcal{C}_{\text{SC-A}} &= \mathcal{C}_{\text{pre\_stage}} + \mathcal{C}_{\text{attn\_stage}} + \mathcal{C}_{\text{post\_stage}}\\
&= 4 \cdot C \cdot H \cdot W + 4 \cdot C \cdot h \cdot w + 6 \cdot H \cdot W \cdot C \\
&+ 2 \cdot (H + W) \cdot (h + w) \cdot (2C).
\end{aligned}
\end{equation}

The computational complexity of the \textbf{Non-local Network} \cite{wang2018non} can be derived by analyzing its key stages: Firstly, feature maps \( \mathbf{A} \) and \( \mathbf{B} \) are projected into embedding spaces (query, key, and value tensors) using \( 1 \times 1 \) convolutions. The computational complexity denotes:
\begin{equation}
\mathcal{C}_{\text{embed}} = 3 \cdot C \cdot H \cdot W + C \cdot h \cdot w.
\end{equation}

Secondly, the combined complexity of the attention map computation and application is:
\begin{equation}
\mathcal{C}_{\text{attn}} = 2 \cdot C \cdot (H \cdot W) \cdot (h \cdot w).
\end{equation}

Finally, the attended features are projected back to the original dimensions of \( \mathbf{A} \) using a \( 1 \times 1 \) convolution, with a complexity of:
\begin{equation}
\mathcal{C}_{\text{proj}} = C \cdot H \cdot W.
\end{equation}

By summing all components, the total computational complexity of the Non-local mechanism is:
\begin{equation}
\begin{aligned}
\mathcal{C}_{\text{Non-local}} &= \mathcal{C}_{\text{embed}} + \mathcal{C}_{\text{attn}} + \mathcal{C}_{\text{proj}}\\
&= 4 \cdot C \cdot H \cdot W + 3 \cdot C \cdot h \cdot w + 2 \cdot C \cdot (H \cdot W) \cdot (h \cdot w).
\end{aligned}
\end{equation}

The \textbf{computational complexity ratio} between SC-A and Non-local is given as shown in Eq. \ref{eq:rationa2}. In practical detection tasks, where \( H \) tends to be large, \textit{the computational complexity of SC-A is significantly lower than that of Non-local. Moreover, this gap becomes increasingly pronounced as \( H \) increases}.
\begin{figure*}[t!]
\begin{equation}
\begin{aligned}
\lambda_{\text{SC-A}} =\frac{\mathcal{C}_{\text{SC-A}}}{\mathcal{C}_{\text{Non-local}}} = 
\frac{10 \cdot H \cdot W + 4 \cdot h \cdot w + 4 \cdot H \cdot h + 4 \cdot H \cdot w + 4 \cdot W \cdot h + 4 \cdot W \cdot w}{4 \cdot H \cdot W + 3 \cdot h \cdot w + 2 \cdot H \cdot W \cdot h \cdot w}
\end{aligned}
\label{eq:rationa2}
\end{equation}
\end{figure*}

Based on Eq.~\ref{eq:rationa2}, the derivation proceeds as follows:
\begin{equation}
\begin{aligned}
\lambda_{\text{SC-A}} &= 
\frac{10 \cdot H \cdot W + 4 \cdot h \cdot w + 4 \cdot (H + W) \cdot (h + w)}{4 \cdot H \cdot W + 3 \cdot h \cdot w + 2 \cdot (H \cdot W) \cdot (h \cdot w)}\\
&\overset{\mathbf{\alpha}}{=}\frac{30 \cdot H^2}{7 \cdot H^2 + 2 \cdot H^4}\\
&=\frac{30}{7 + 2 \cdot H^2}
\label{eq:derive1}
\end{aligned}
\end{equation}
where, Strategy \(\mathbf{\alpha}\) aligns with {Assumption} \ref{assu4:condition} and {Assumption} \ref{assu5:condition}.

\section{Materials \& Experiments}
\label{sec: experiment}

\subsection{Experimental setting}

\begin{figure} [t!]
	\centering
            \subfloat[\small Surveillance images]{
		\includegraphics[width=2.5cm,height=2.3cm]{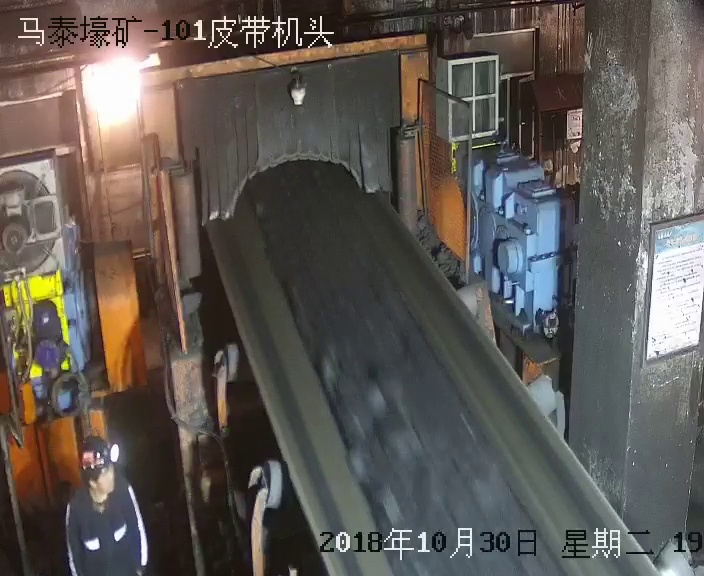}}
        \subfloat[\small Online images]{
		\includegraphics[width=2.5cm,height=2.3cm]{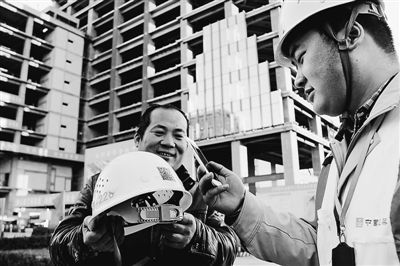}}
        \subfloat[\small Artificial images]{
		\includegraphics[width=2.5cm,height=2.3cm]{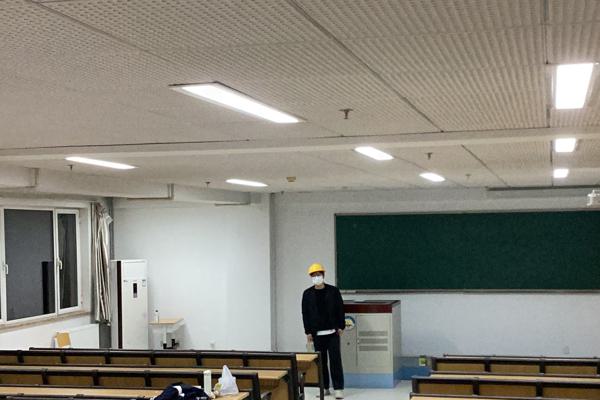}}
\caption{The example of HBDD. There are example of imges captured by real coal mine surveillance system, online crawler technology and simulated real scenes.}
\label{HBDD}
\end{figure}

\textit{Datasets $\&$ Models}. We evaluate the performance of ED-TOOLBOX and its enhancement on detection models using public datasets, including ImageNet \cite{deng2009imagenet} (for classification tasks) and COCO-2017 \cite{lin2014microsoft} (for detection tasks). Notably, we contribute a new dataset, the \textit{Helmet Band Detection Dataset\footnote{\url{https://github.com/word-ky/Edge-TOOLBOX/tree/main}}} (HBDD), as shown in Fig. \ref{HBDD}. It consists of 8100 images, each with a resolution of 640x640, with an 8:2 split for training and testing. It contains four categories: \textit{nohelmet} (Non-Wearing a helmet), \textit{noband} (Wearing a helmet without a hatband), \textit{head} (Worker's head) and \textit{helmet} (Wearing a helmet properly). This dataset includes real surveillance images, web-crawled images, and simulation-generated images. Due to strict supervision in practical surveillance, real images rarely include instances of workers improperly wearing helmets, therefore, we simulate improper helmet usage to supplement the negative samples, as shown in Fig \ref{HBDD} (c). For classification tasks, we use ResNet-50 \cite{he2016deep} and ViT-s/16 \cite{wu2020visual} as backbone models. In detection tasks, we use the popular YOLOv8 \cite{ali2024yolo} (which is more stable in practical applications compared to the latest v11 \cite{khanam2024yolov11}) and the traditional SSD  \cite{liu2016ssd} as backbone models.


\textit{Devices $\&$ Implement details}. In the \textit{Visual Surveillance Simulation Experiment} \ref{sec:Simulation}, we use a terminal computing device with an NVIDIA Jetson TX2 and a Tencent Cloud Server\footnote{\url{https://www.tencentcloud.com/}} with two RTX 4090 GPU to simulate the \textit{Internet of Video Things} (IoVT) system, respectively. Aditionally, the system’s data transfer rate is set to 100Mbps, and the communication environment has an uplink bandwidth of 106.4Mbps, downlink bandwidth of 41.3Mbps, and a Ping of 11ms. The detailed experimental environment and simulation setup are shown in TABLE \ref{table:setup}. Other all experiments are conducted on the two RTX 4090.   

\begin{figure*} [t]
\centering
\includegraphics[width=0.7\textwidth,height=0.65\textwidth]{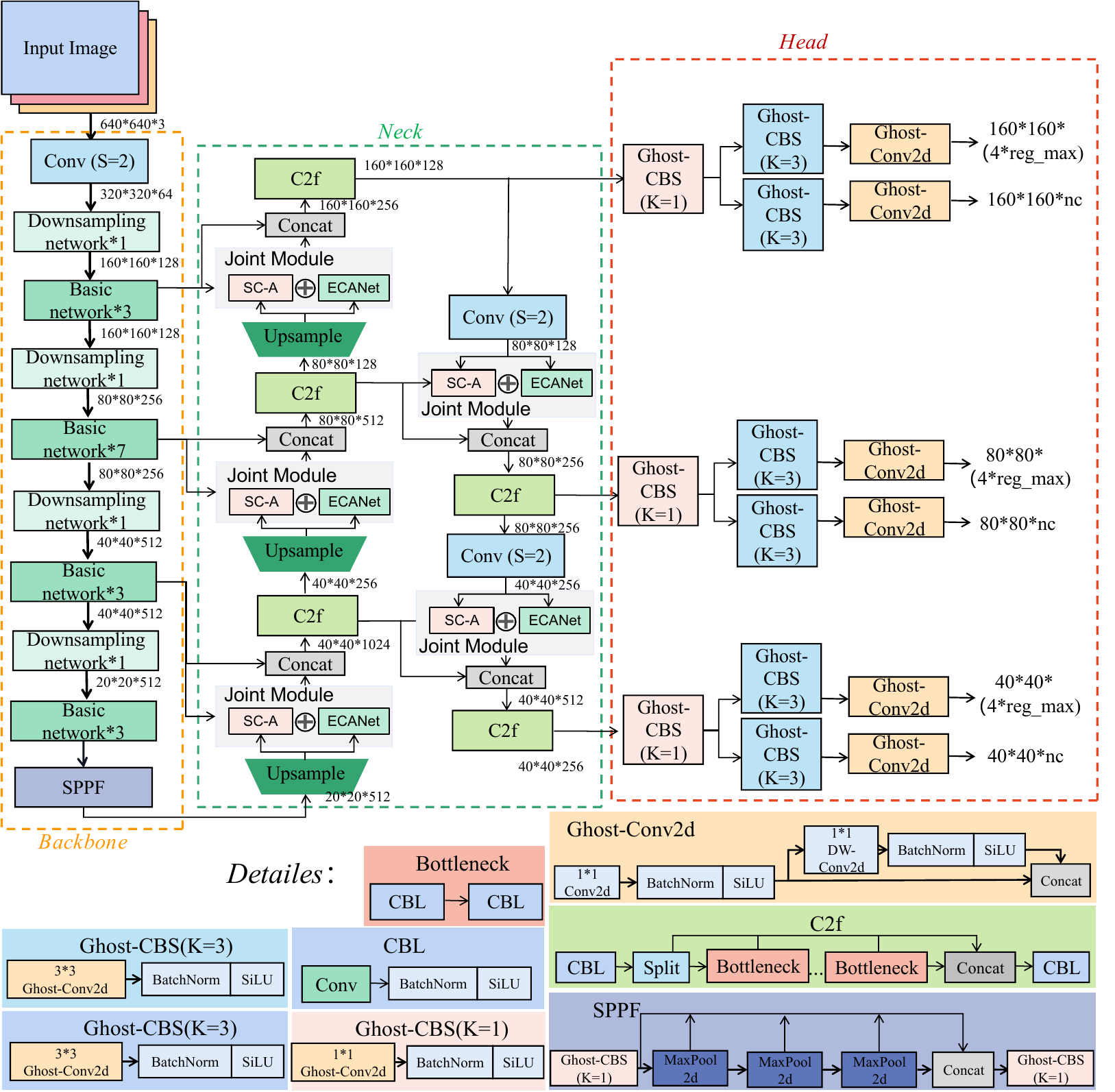}
\caption{Structure of ED-YOLO. Based on the YOLO backbone model, YOLOv8-s, ED-TOOLBOX implements improvements for edge deployment. Specifically, Rep-DConvNet (as shown in Fig. \ref{fig:rep}) is used to enhance the Backbone, Joint Modules (as shown in Fig. \ref{fig:SCA}) are inserted into the Neck to connect upstream and downstream modules, and the Efficient Head (as shown in Fig. \ref{fig:overview}) replaces the original Head.}
\label{fig:YOLODBtoolbox}
\end{figure*}

The complete architecture of the detection model based on ED-TOOLBOX is shown in Fig. \ref{fig:YOLODBtoolbox}, as discussed in \textit{Sections} \ref{sec:dection}, \ref{sec:Ablation}, and \ref{sec:Simulation}. Moreover, in \textit{Section} \ref{sec:effectiveness}, for Rep-DConvNet, we construct the full network architecture based on the setup described in the study by \cite{zhang2018shufflenet}. For the integration of the SC-A module with backbone models, we implement it following the approach outlined in \cite{huang2025generic}.

\begin{table}[h!]
\centering
\caption{The detailed experimental setups.}
\begin{tabular}{ccc}
\hline
\multirow{8}{*}{\textbf{Devices}} & \multirow{4}{*}{Embedded Devices} & AI Performance: 1.33 TFLOPS                          \\
                                  &                                  & GPU: 256-core NVIDIA GPU                        \\
                                  &                                  & GPU Max Frequency: 1.3 GHz                        \\
                                  &                                  & CPU: Dual-core NVIDIA CPU                        \\ \cline{2-3} 
                                  & \multirow{4}{*}{Servers}         & CPU: AMD EPYC 7453       \\
                                  &                                  & RAM: 64.4GB              \\
                                  &                                  & SSD: 451.0GB             \\
                                  &                                  & GPU: RTX 4090            \\ \hline
                                  \multirow{4}{*}{\textbf{Model Training}} & Batch size    & 20      \\
                                        & Epochs        & 150 \\ 
                                        & Learning rate & 0.02  \\ 
                                        & Optimizer     & Adam  \\
                                       \hline
\multicolumn{2}{c}{\multirow{3}{*}{\textbf{Software Tools}}} & CUDA 11.3 \& cuDNN 8.2.0 \\
\multicolumn{2}{c}{}                                                         & PyTorch 2.1                 \\
\multicolumn{2}{c}{}                                                         & Python 3.7               \\ \hline
\end{tabular}
\label{table:setup}
\end{table}

\textit{Evaluate metrics $\&$ Experimental goals}. We evaluate the effectiveness of SC-A and Rep-DConvNet in classification tasks using accuracy. For detection tasks, we use the mean Average Precision (mAP), specifically \textit{mAP@IoU}=0.5, which calculates the model's average precision at an \textit{Intersection over Union} (IoU) threshold of 0.5 to assess the performance of ED-TOOLBOX-assisted models. Computational complexity (FLOPs/G) and parameter count (Paras/M) are used to evaluate model size. Furthermore, we assess the real-time performance of the detection models using throughput (\textit{Frames Per Second}, FPS) and total throughput (\textit{Total Frames Per Second}, TFPS), where TFPS includes both model inference time and data transmission time within the system. This metric reflects the actual latency taken for the detection function to respond to user requests in the Internet of Video Things (IoVT) system. 

Additionally, all experiments are designed to address the following research questions: \begin{itemize} 
    \item \textbf{RQ1 (Resource Utilization):} Do the components of ED-TOOLBOX exhibit a lightweight scale? Can ED-TOOLBOX-assisted detection models (as shown in Fig. \ref{fig:YOLODBtoolbox}) be effectively adapted to resource-constrained edge devices?
    \item \textbf{RQ2 (Real-time Performance):} Can the ED-TOOLBOX-assisted detection models achieve real-time inference? 
    \item \textbf{RQ3 (Accuracy):} Can the components of ED-TOOLBOX achieve high-accuracy classification? Can our edge detection models (as shown in Fig. \ref{fig:YOLODBtoolbox}) perform accurate object detection? 
    \item \textbf{RQ4 (Adaptability):} Can the components of ED-TOOLBOX assist different models in achieving high performance? 
    \item \textbf{RQ5 (Practical Application):} Can the ED-TOOLBOX-assisted detection models achieve precise and real-time object detection, particularly for small objects, in real-world scenarios? 
\end{itemize}

\textit{Experiment Process}. The experiments consist of ``\textit{Effectiveness Evaluation}", ``\textit{Detection Comparative Experiment}", ``\textit{Ablation Study}", and ``\textit{Visual Surveillance Simulation Experiment}". Below is the detailed experimental procedure.


\textbf{Effectiveness Evaluation.} We perform classification tasks on the ImageNet dataset using the RTX 4090 to evaluate the effectiveness of Rep-DConvNet and SC-A. Specifically, for Rep-DConvNet, we use nine state-of-the-art (SOTA) lightweight networks, including MobileNetV1 \cite{howard2017mobilenets} / V2 \cite{sandler2018mobilenetv2}, ShuffleNetV1 \cite{zhang2018shufflenet} / V2 \cite{ma2018shufflenet}, GhostNet \cite{han2020ghostnet}, MobileNeXt \cite{zhou2020rethinking}, and PP-LCNet \cite{cui2021pp}, to evaluate its resource utilization (for \textbf{RQ1}) and classification accuracy (for \textbf{RQ3}). For SC-A, we use seven SOTA attention networks: Non-local \cite{wang2018non}, MOATransformer \cite{patel2022aggregating}, CrossFormer \cite{zhang2023crossformer}, DAT Attention \cite{xia2022vision}, and Pyramid Split Attention \cite{zhang2021epsanet} (PSA) to evaluate its accuracy improvement (for \textbf{RQ3}) and model-scale changes (for \textbf{RQ1}). Additionally, we use ResNet-50 \cite{he2016deep} and ViT-s/16 \cite{wu2020visual} as backbone models to assess the plug-and-play capability of SC-A (for \textbf{RQ4}).


\textbf{Detection Comparative Experiment.} We build edge detection models based on ED-TOOLBOX. To evaluate performance differences, we compare our edge detection model against seven state-of-the-art (SOTA) methods and five variants of YOLOv8, using both the public COCO dataset and our proprietary HBDD dataset on the two RTX 4090. The evaluation focuses on detection accuracy (for \textbf{RQ3}), model scale (for \textbf{RQ1}), and inference real-time performance (for \textbf{RQ2}). Specifically, we compare large models such as DETR-101 (DC5-ResNet-101) \cite{End-to-end}, YOLOv7-X \cite{YOLOv7}, YOLOv11-X \cite{khanam2024yolov11}, FCOS-101 (ResNet-101) \cite{FCOS}, and YOLOv8-x/l/m, with lightweight models like YOLOv7-tiny-SiLU \cite{YOLOv7}, SSD-VGG \cite{liu2016ssd}, PPYOLOEs \cite{PP-YOLOE}, YOLOv5-s, and YOLOv8-s/n. Additionally, to evaluate the plug-and-play capability of ED-TOOLBOX for detection models (for \textbf{RQ4}), we build ED-YOLO by using YOLOv8-s as the backbone model (as illustrated in Fig. \ref{fig:YOLODBtoolbox}), and also integrate Rep-DConvNet into the feature extraction network of the traditional SSD-VGG \cite{liu2016ssd}, resulting in ED-SSD.


\textbf{Ablation Study.} Unlike the effectiveness experiments, the ablation experiment is designed to systematically evaluate the contribution of each component in ED-TOOLBOX to the performance enhancement of edge detection models. We use the HBDD dataset and conduct the experiments on two RTX 4090 platform. Specifically, we objectively assess the performance gains of each component using detection accuracy (for \textbf{RQ3}), model scale (for \textbf{RQ1}), and inference real-time performance (for \textbf{RQ2}). Additionally, we showcase the feature extraction capability of the Rep-DConvNet-based YOLO backbone network using feature maps, and use heatmaps to subjectively assess the enhancement in small object detection performance provided by SC-A.


\textbf{Visual Surveillance Simulation Experiment.} Based on the experimental setup (as detailed in ``\textit{Devices \& Implementation Details}"), we construct an IoVT system to simulate a real-world surveillance scenario and perform the helmet band detection task to evaluate the practical application effect of our ED-TOOLBOX. Specifically, we deploy ED-YOLO on both cloud servers and edge devices to assess detection accuracy and real-time performance. The experimental results will demonstrate the practical value of ED-TOOLBOX in real-world applications (for \textbf{RQ5}).

\subsection{Effectiveness Evaluation}
\label{sec:effectiveness}
We conduct an Effectiveness Evaluation to assess the performance of the key components in ED-TOOLBOX, namely Rep-DConvNet and the SC-A network, as shown in TABLE \ref{tab:effectiveness-rep} and TABLE \ref{tab:SC-A}. Additional subjective experimental results are provided in \textit{Section} \ref{sec:dection}.

\begin{table}[h!]
\setlength{\tabcolsep}{0.3cm}
\centering
\caption{Comparison of Lightweight Models on ImageNet. Our Rep-DConvNet demonstrates the ability to achieve both \textit{low computational resource usage and accuracy improvement}. ''Acc'' refers to classification accuracy. \textcolor{red}{The best results} are marked in red, and \textcolor{blue}{the second-best results} are marked in blue.}
\begin{tabular}{lccc}
\hline
\textbf{Model}        & \textbf{\#Param/M}$\downarrow$ & \textbf{\#FLOPs/M}$\downarrow$ & \textbf{Acc/\%}$\uparrow$ \\ \hline
MobileNetV1 \cite{howard2017mobilenets}   & \textcolor{blue}{2.6}                  & 333.0                & 70.5            \\
MobileNetV2 \cite{sandler2018mobilenetv2}   & 3.5                  & 314.2                & 72.2            \\
ShuffleNetV1 \cite{zhang2018shufflenet} & 5.5                 & 536.0                & 72.4            \\
ShuffleNetV2 \cite{ma2018shufflenet} & 3.5                  & 304.5                & 71.7            \\
GhostNet \cite{han2020ghostnet}         & 5.2                  & \textcolor{red}{150.0}               & 71.1            \\
MobileNeXt \cite{zhou2020rethinking}     & 3.5                 & 351.7               & \textcolor{red}{74.2}            \\
PP-LCNet \cite{cui2021pp}         & 3.0                  & 168.4               & 71.7          \\ 
\cellcolor{gray!30}Rep-DConvNet        &      \cellcolor{gray!30}\textcolor{red}{1.8}             &    \cellcolor{gray!30}\textcolor{blue}{154.9}             &     \cellcolor{gray!30}\textcolor{blue}{72.7}       \\ \hline
\end{tabular}
\label{tab:effectiveness-rep}
\end{table}

Rep-DConvNet demonstrates significant advantages across three key metrics. Specifically, its parameter count (\#Param) is only 1.8M, which is notably lower than other leading lightweight models, reducing the parameter count by 30.8\% compared to the second best, MobileNetV1 (2.6M). Furthermore, Rep-DConvNet's computational complexity (\#FLOPs) is 154.9M, slightly higher than that of GhostNet (150.0M), but still maintains competitive computational efficiency compared to other models, indicating its high resource efficiency. Additionally, Rep-DConvNet achieves an accuracy (Acc) of 72.7\%, only trailing MobileNeXt (74.2\%), demonstrating that it effectively enhances model performance while maintaining low computational and parameter overhead, thus achieving an optimal balance between accuracy and efficiency.

\begin{table}[h]
\setlength{\tabcolsep}{0.01cm}
\centering
\caption{Comparison of Attention Networks on ImageNet. Our SC-A demonstrates the ability to achieve \textit{low resource utilization while effectively improving accuracy}. ''ORG'' refers to the original backbone model, ''Acc'' denotes classification accuracy, and ''$\Delta$Acc'' indicates the accuracy improvement. \textcolor{red}{The best results} are marked in red, and \textcolor{blue}{the second-best results} are marked in blue.
}
\begin{tabular}{lccccc}
\hline
\textbf{Backbone} & \textbf{Model}        & \textbf{\#Param/M}$\downarrow$ & \textbf{\#FLOPs/G}$\downarrow$ & \textbf{Acc/\%}$\uparrow$ & \textbf{$\Delta$Acc/\%}$\uparrow$ \\ \hline
\multirow{7}{*}{ResNet50} 
& ORG \cite{he2016deep}   & 25.5                  & 195.2                & 75.3            & --          \\
 & Non-local \cite{wang2018non}   & \textcolor{blue}{26.6}                 & 2825.9               & \textcolor{red}{79.6}            & \textcolor{red}{+4.3}          \\
 &  MOATransformer \cite{patel2022aggregating}   & 39.6                 & 199.9               & 76.7           & {+1.4}          \\
 & CrossFormer \cite{zhang2023crossformer}     & 33.4                 & \textcolor{blue}{196.5}               & 77.0           & {+1.7}          \\
 & DAT \cite{xia2022vision}        & 53.8                 & 199.8               & \textcolor{blue}{78.4}            & \textcolor{blue}{+3.1}          \\
  & PSA \cite{zhang2021epsanet} & 33.9                 & 212.4               & 77.8            & {+2.5}          \\
 & \cellcolor{gray!30} SC-A                     & \cellcolor{gray!30}\textcolor{red}{25.5}                & \cellcolor{gray!30}\textcolor{red}{195.9}               & \cellcolor{gray!30}78.2              &\cellcolor{gray!30} {+2.9}          \\ \hline\hline
\multirow{7}{*}{ViT-s/16} 
& ORG \cite{wu2020visual}   & 30.0                  & 4.3                & 78.6            & --          \\
 & Non-local \cite{wang2018non}   & \textcolor{blue}{31.1}                 & 2635.0             & \textcolor{red}{81.5}            & \textcolor{red}{+2.9}          \\
 &  MOATransformer \cite{patel2022aggregating}   & 44.1                 & 9.0                & 79.5            & {+0.9}          \\
 & CrossFormer \cite{zhang2023crossformer}     & 37.9                 & \textcolor{blue}{5.6}                & 80.5            & {+1.9}          \\
 & DAT \cite{xia2022vision}          & 58.3                 & 8.9                & 80.1            & {+1.5}          \\
  & PSA \cite{zhang2021epsanet} & 38.4                 & 21.5               & 79.5            & {+0.9}          \\
 & \cellcolor{gray!30} SC-A                     & \cellcolor{gray!30} \textcolor{red}{30.0}                 & \cellcolor{gray!30}\textcolor{red}{5.0}                &\cellcolor{gray!30} \textcolor{blue}{80.8}              &\cellcolor{gray!30} \textcolor{blue}{+2.2}          \\ \hline
\end{tabular}
\label{tab:SC-A}
\end{table}

\begin{figure*} [t!]
	\centering

             \subfloat[DETR-101]{
		\includegraphics[width=2.5cm,height=2.3cm]{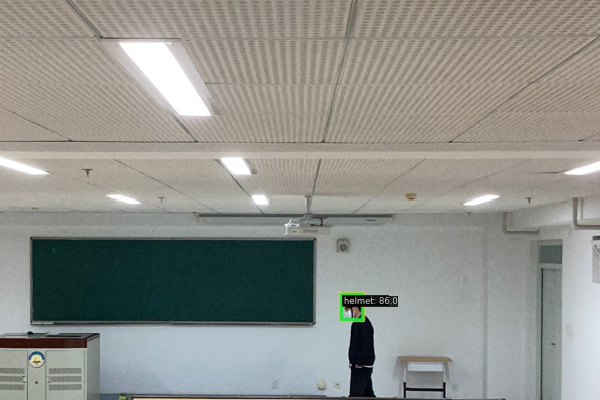}}
         \hspace{-3mm}
         \subfloat[FCOS-101]{
		\includegraphics[width=2.5cm,height=2.3cm]{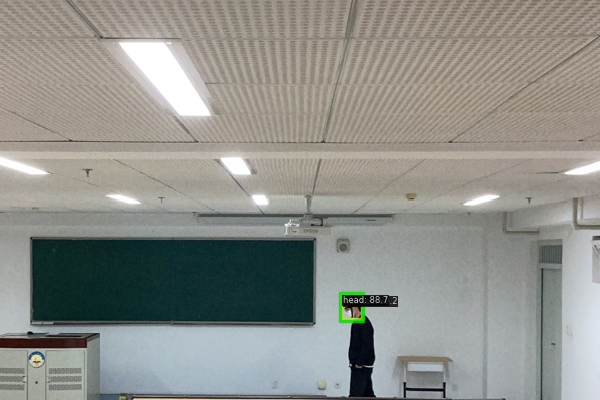}}
         \hspace{-3mm}
            \subfloat[YOLOv7-x]{
		\includegraphics[width=2.5cm,height=2.3cm]{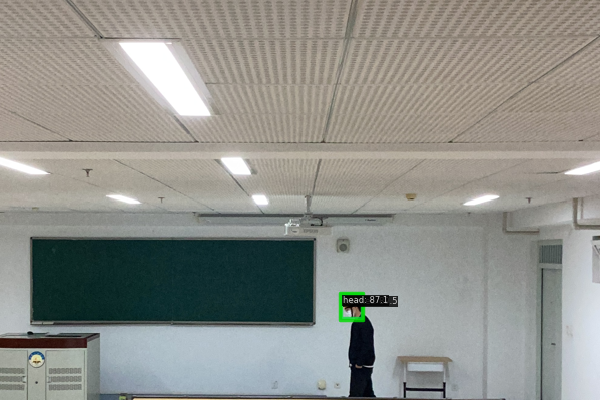}}
  \hspace{-3mm}
   \subfloat[YOLOv7-tiny-SiLU]{
		\includegraphics[width=2.5cm,height=2.3cm]{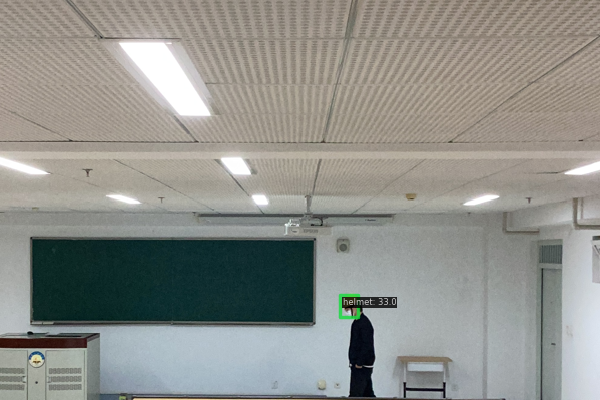}}
         \hspace{-3mm}
             \subfloat[PPYOLOEs]{
		\includegraphics[width=2.5cm,height=2.3cm]{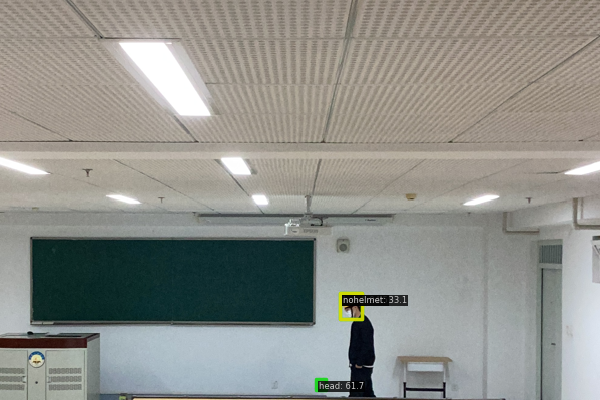}}
         \hspace{-3mm}
         \subfloat[YOLOv8-s]{
		\includegraphics[width=2.5cm,height=2.3cm]{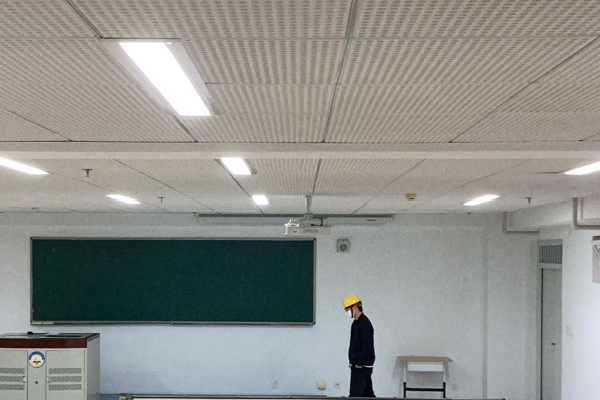}}
         \hspace{-3mm}
            \subfloat[ED-YOLO]{
		\includegraphics[width=2.5cm,height=2.3cm]{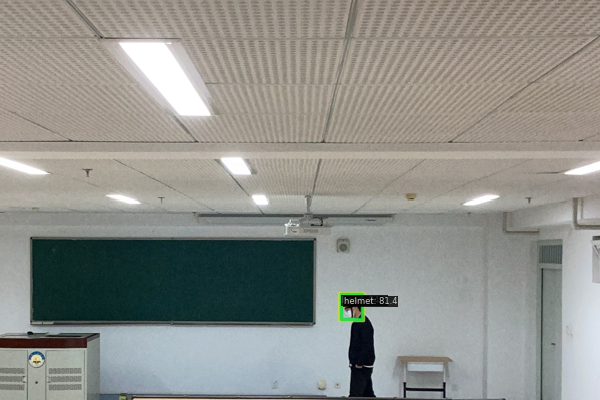}}
  \hspace{-3mm}\\
  \centering{HBDD}\\
     
            \subfloat[DETR-101]{
		\includegraphics[width=2.5cm,height=2.3cm]{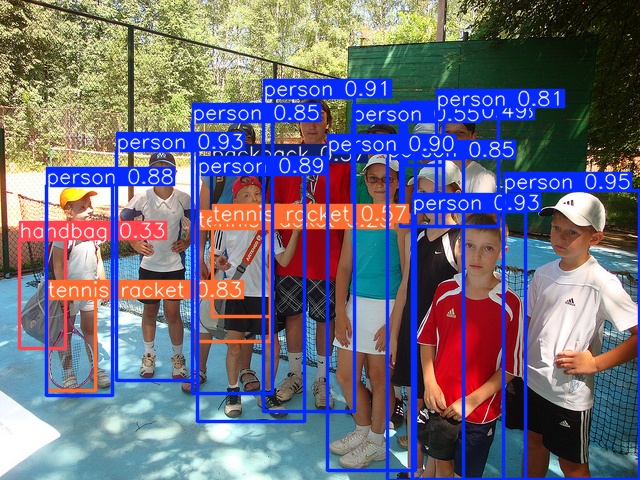}}
         \hspace{-3mm}
         \subfloat[FCOS-101]{
		\includegraphics[width=2.5cm,height=2.3cm]{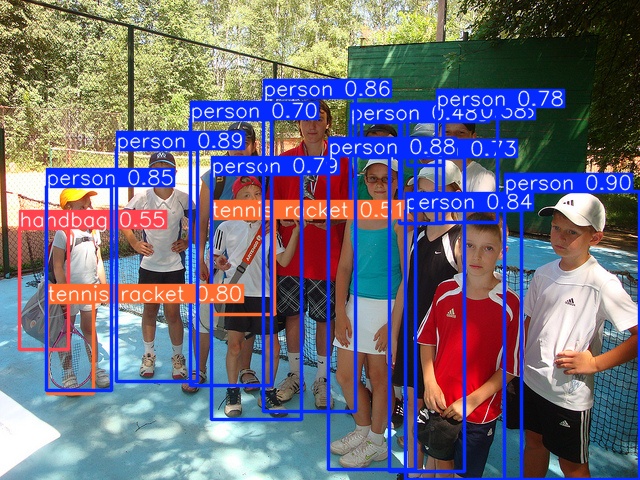}}
         \hspace{-3mm}
            \subfloat[YOLOv7-x]{
		\includegraphics[width=2.5cm,height=2.3cm]{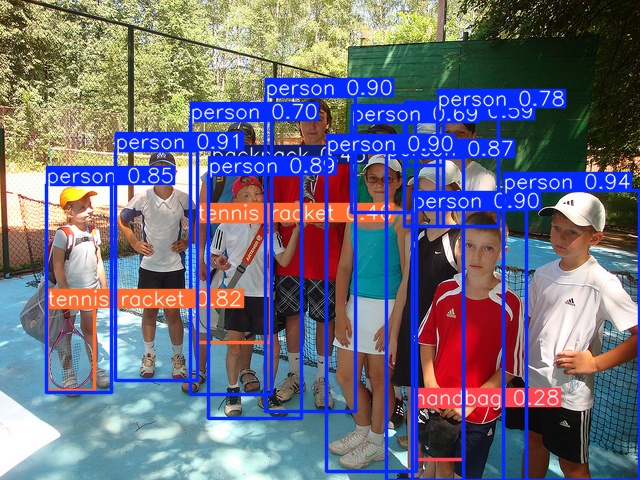}}
  \hspace{-3mm}
   \subfloat[YOLOv7-tiny-SiLU]{
		\includegraphics[width=2.5cm,height=2.3cm]{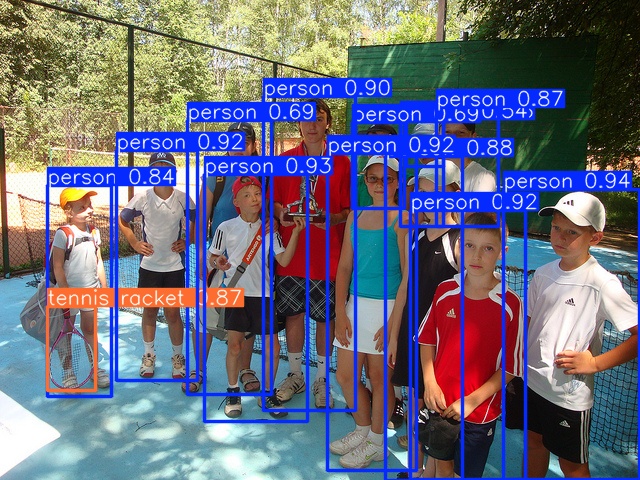}}
         \hspace{-3mm}
             \subfloat[PPYOLOEs]{
		\includegraphics[width=2.5cm,height=2.3cm]{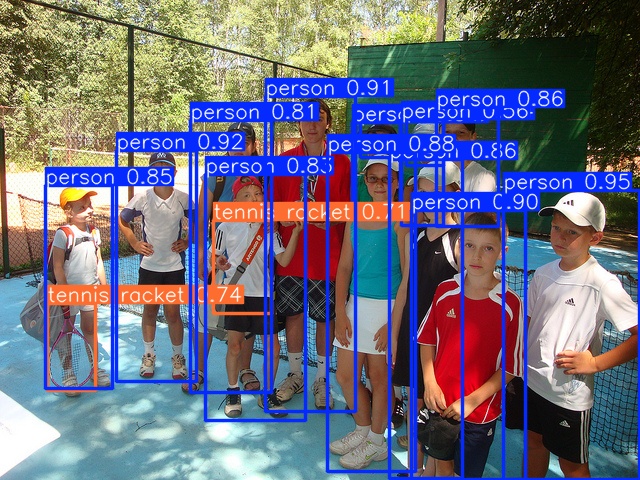}}
         \hspace{-3mm}
         \subfloat[YOLOv8-s]{
		\includegraphics[width=2.5cm,height=2.3cm]{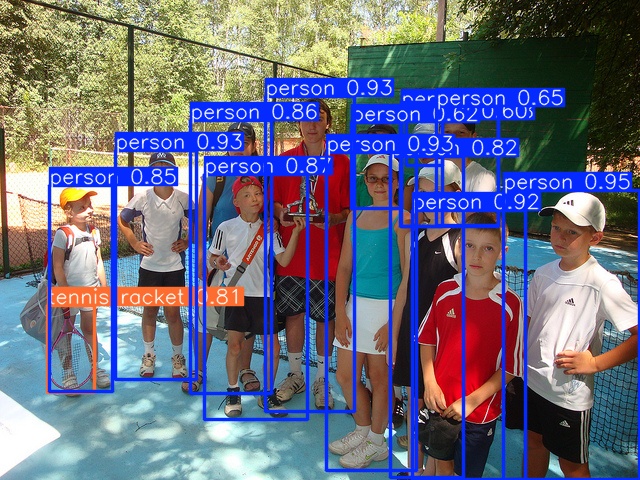}}
         \hspace{-3mm}
            \subfloat[ED-YOLO]{
		\includegraphics[width=2.5cm,height=2.3cm]{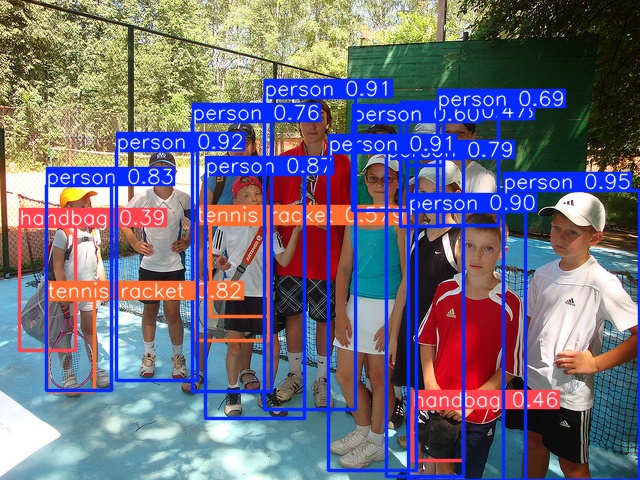}}
  \hspace{-3mm}\\
  \subfloat[DETR-101]{
		\includegraphics[width=2.5cm,height=2.3cm]{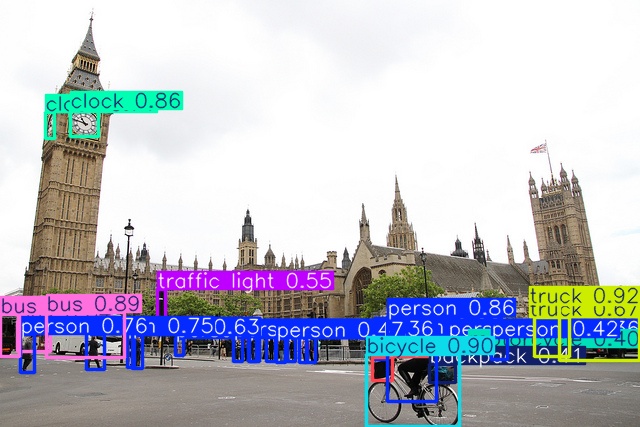}}
         \hspace{-3mm}
         \subfloat[FCOS-101]{
		\includegraphics[width=2.5cm,height=2.3cm]{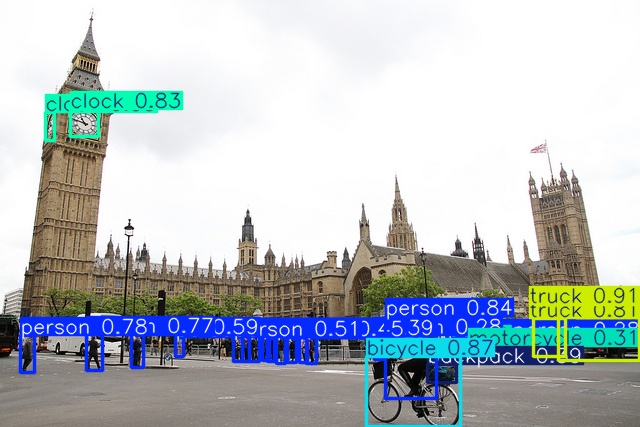}}
         \hspace{-3mm}
            \subfloat[YOLOv7-x]{
		\includegraphics[width=2.5cm,height=2.3cm]{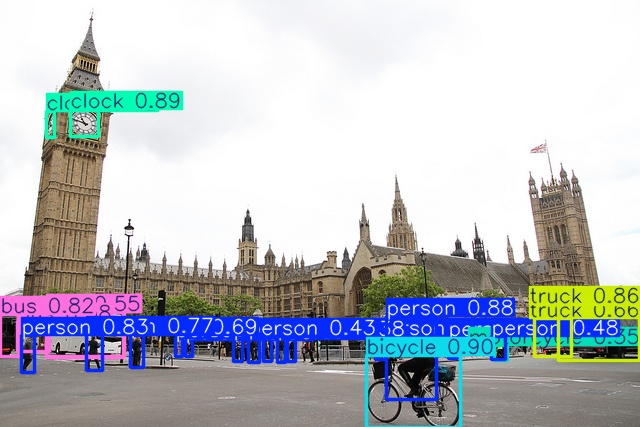}}
  \hspace{-3mm}
   \subfloat[YOLOv7-tiny-SiLU]{
		\includegraphics[width=2.5cm,height=2.3cm]{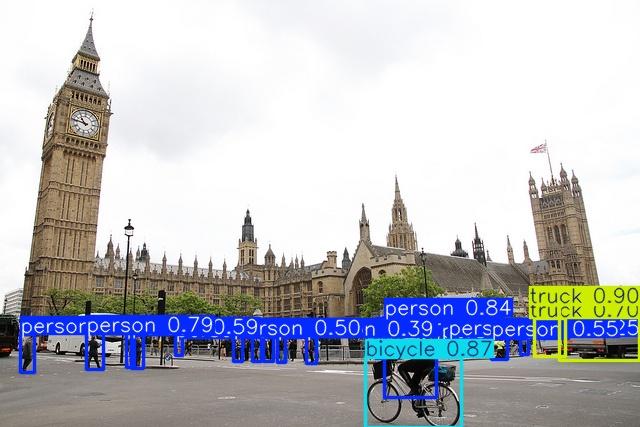}}
         \hspace{-3mm}
             \subfloat[PPYOLOEs]{
		\includegraphics[width=2.5cm,height=2.3cm]{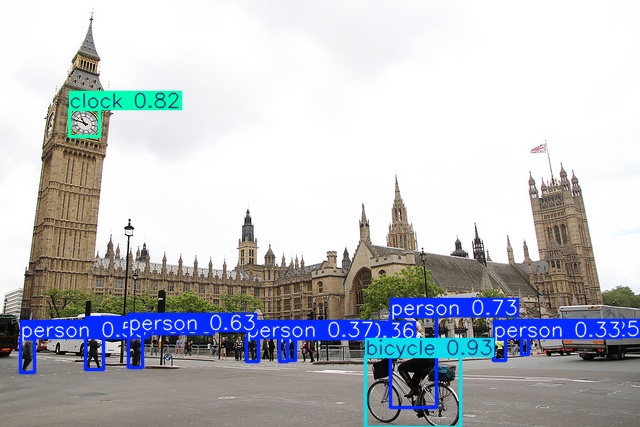}}
         \hspace{-3mm}
         \subfloat[YOLOv8-s]{
		\includegraphics[width=2.5cm,height=2.3cm]{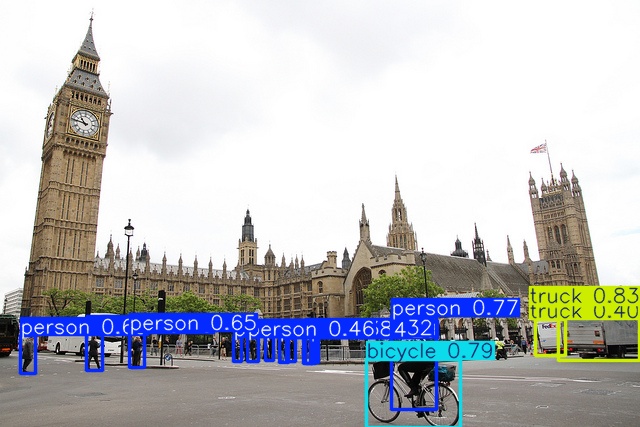}}
         \hspace{-3mm}
            \subfloat[ED-YOLO]{
		\includegraphics[width=2.5cm,height=2.3cm]{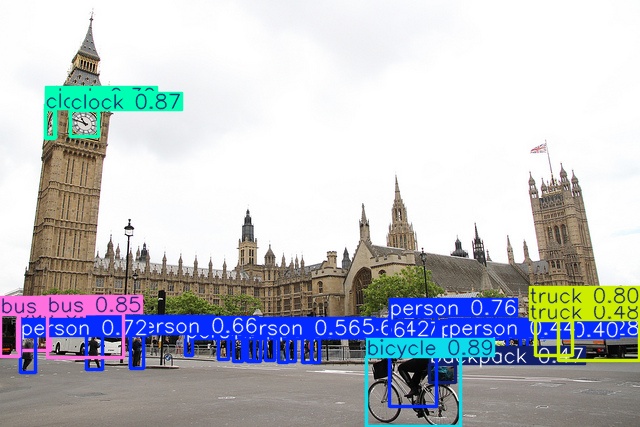}}
  \hspace{-3mm}\\
  \centering{COCO}
          
  \caption{Comparison of detection results on our HBDD and the public COCO datasets. Lightweight models such as YOLOv7-tiny-SiLU, PPYOLOEs, and YOLOv5-s exhibit significant false detection and missed detections, such as (\textit{f}), (\textit{m}), and (\textit{r}). In contrast, \textit{our model achieves detection performance comparable to that of more complex models}, effectively detecting small-sized objects (helmet band in (\textit{g})) and occluded objects (person in (\textit{u})).}
  \label{fig:coco&hbdd}
\end{figure*}

TABLE \ref{tab:SC-A} demonstrates the exceptional performance of SC-A across all metrics, with particularly notable advantages in terms of parameter count and computational complexity. Specifically, SC-A maintains zero increase in parameter count across both backbone models, while only exhibiting a minimal increase in computational complexity, highlighting its efficiency in resource utilization. Moreover, SC-A achieves an accuracy improvement of 2.9\% on the ResNet50 backbone, and +2.2\% on the ViT-s/16 backbone, second only to the Non-local method, further validating its effectiveness in enhancing classification accuracy. Overall, SC-A demonstrates outstanding adaptability and efficiency across different backbone networks, showcasing the optimal balance between accuracy improvement and computational resource optimization.

\subsection{Detection Comparative Experiment}
\label{sec:dection}


To evaluate the benefits of ED-TOOLBOX on detection models, we build the ED-YOLO and ED-SSD models using YOLOv8-s and SSD-VGG, respectively, as examples. These models are compared with state-of-the-art methods on the public COCO dataset and our custom HBDD dataset, focusing on computational resource usage, real-time performance, and detection accuracy. The visual detection results are presented in Fig. \ref{fig:coco&hbdd}, while TABLE \ref{tab:detection-comparison} provides objective numerical comparison on the HBDD. Additionally, we report the GPU memory utilization during training to further assess the computational efficiency of the models.

\begin{table}[t]
\setlength{\tabcolsep}{0.2cm}
\centering
    \caption{Comparison of model detection performance on high-performance devices using the HBDD. ED-TOOLBOX significantly enhances the accuracy of the original models and effectively reduces resource usage. For example, ED-YOLO achieves the highest accuracy among lightweight models while maintaining lower resource utilization. \textcolor{red}{The best results} are marked in red, and \textcolor{blue}{the second-best results} are marked in blue. In the \textbf{lightweight models, the best mAP} is highlighted in bold.}
\begin{tabular}{cccccc}
\hline
\textbf{Model}        & \textbf{mAP/\%}$\uparrow$ & \textbf{FPS}$\uparrow$   & \textbf{Param/M}$\downarrow$ & \textbf{FLOPs/G}$\downarrow$\\ \hline
DETR-101\cite{End-to-end}         & 95.47  & 45.4  & 60.1    & 253.8   \\
YOLOv7-x \cite{YOLOv7}        & 96.13  & 58.6  & 75.6    & 192.3   \\
FCOS-101  \cite{FCOS}       & 95.53  & 70.8  & 52.3    & 132.7   \\
YOLOv11-x  \cite{khanam2024yolov11}       & \textcolor{blue}{98.17}   &  57.1  &   56.9   &  196.0   \\
YOLOv7-tiny \cite{YOLOv7}     & 78.45  & {93.3} & {13.8}     & 20.5    \\
 PPYOLOE-s \cite{PP-YOLOE} &  83.35  &  92.1  &  15.4     &  23.1    \\
 YOLOv5-s   &  80.19  &  {92.7} &  13.1     &  21.5    \\
\hdashline 
YOLOv8-x          & \textcolor{red}{98.24}  & 52.1 & 61.5   & 227.9  \\
YOLOv8-l          & 96.73  & 65.2 & 47.1   & 163.8   \\
YOLOv8-m          & 90.62  & 79.0 & 24.7   & 82.6   \\
 YOLOv8-s  &  82.15  &  91.4 &  14.3   &  25.8   \\
 YOLOv8-n  &  69.32  &  \textcolor{red}{97.5} &  \textcolor{red}{5.9}    &  \textcolor{red}{9.6}    \\ 
SSD-VGG \cite{liu2016ssd}         &  74.29  &  77.2 &   36.0    &   98.6   \\\hdashline
\rowcolor{gray!30} ED-SSD   & 83.51  &79.4 & 27.6   & 82.3  \\
\rowcolor{gray!30} ED-YOLO  &  \textbf{91.34}  & \textcolor{blue}{93.8} & \textcolor{blue}{10.9}   & \textcolor{blue}{19.7}  \\ \hline
\end{tabular}
\label{tab:detection-comparison}
\end{table}

\begin{figure} [b!]
	\centering
       \subfloat{	
        \includegraphics[width=0.4\textwidth
        ]{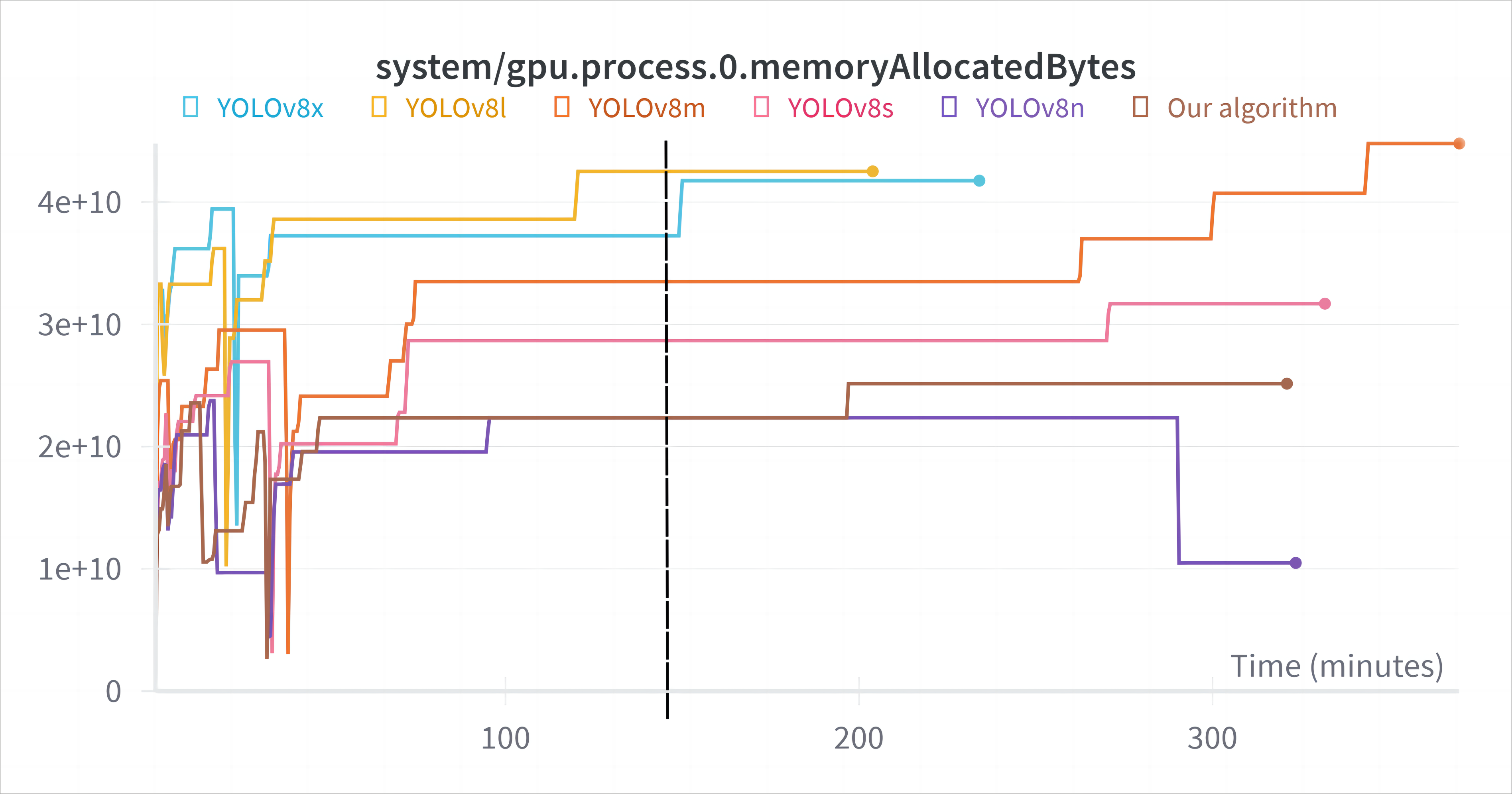}
        
}
        
  \caption{Compare the resource utilization of ED-YOLO with the n/s/m/l/x versions of the original YOLOv8 during the training phase. The figure shows that in the stable phase, \textit{the GPU memory usage of ED-YOLO is only slightly higher than that of YOLOv8n and is even comparable at certain stages}, as indicated by the \textit{black dashed line}. }
  \label{fig:gpu}
  
\end{figure}

\textbf{Numerical Results.} TABLE \ref{tab:detection-comparison} presents the experimental results of different models on our HBDD dataset. Specifically, ED-YOLO achieves the highest detection accuracy among lightweight models (91.34\% mAP), while reducing parameters by 17\% and computational complexity (FLOPs) by 23\% compared to the backbone model, YOLOv8-s. The parameter count and computational complexity of ED-YOLO are only slightly higher than YOLOv8-n, but much lower than that of larger models. Furthermore, ED-SSD improves accuracy by 9.22\% while reducing FLOPs by 16\%, compared to the original ED-SSD. These results demonstrate that ED-TOOLBOX effectively addresses the trade-off between model size and performance, making it particularly well-suited for edge detection tasks with lower resource consumption. Additionally, we report the performance of several typical models on the COCO2017 dataset, such as FCOS-101: 65.4\%, YOLOv7-x: 69.8\%, YOLOv11-x: 73.3\%, YOLOv8-x: 71.6\%, PPYOLOEs: 58.7\%, YOLOv7-tiny-SiLU: 54.3\%, YOLOv8-s: 57.9\%, and ED-YOLO: 61.5\%. From these results, we can observe that, compared to other models, \textit{the ED-TOOLBOX-enhanced models achieve detection accuracy similar to that of more complex models while maintaining lower resource utilization.  These results also highlight the adaptability of ED-TOOLBOX across different backbone models}.


\begin{figure*} [t!]
	\centering
            \subfloat[Input Image]{
		\includegraphics[width=3.3cm,height=2.4cm]{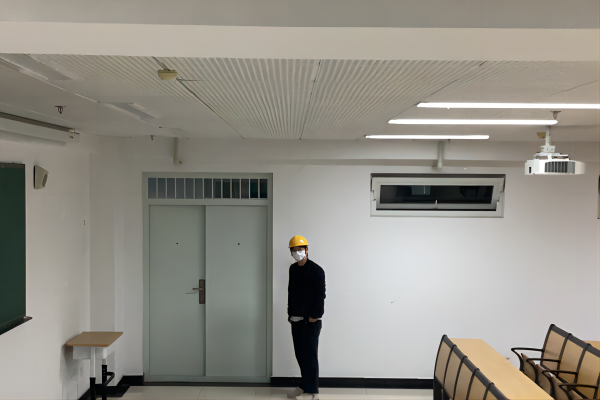}}
         \subfloat[\centering The shallow layer of our Backbone]{
		\includegraphics[width=3.3cm,height=2.4cm]{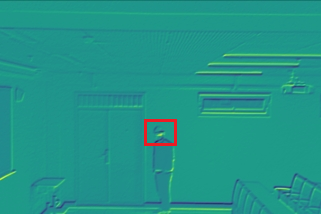}}
         \subfloat[\centering The shallow layer of original DarkNet-53]{
		\includegraphics[width=3.3cm,height=2.4cm]{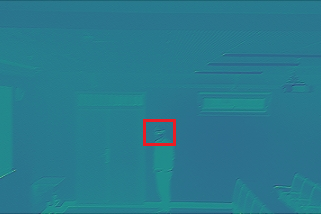}}
         \subfloat[\centering The deep layer of our Backbone]{
		\includegraphics[width=3.3cm,height=2.4cm]{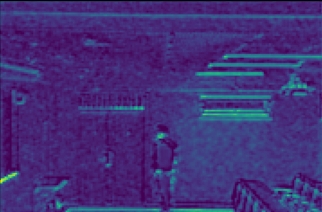}}
         \subfloat[\centering The deep layer of original DarkNet-53]{
		\includegraphics[width=3.3cm,height=2.4cm]{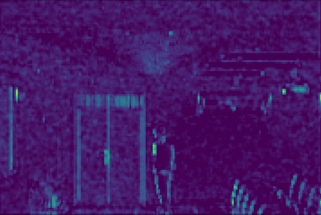}}
         \\
  \caption{Comparison of feature extraction effects. Compared to the original DarkNet-53, our Backbone module (as shown in Fig. \ref{fig:YOLODBtoolbox}) \textit{extracts more detailed features in its shallow layers}, such as the helmet marked by the red box in (\textit{b}), while \textit{its deep layers effectively capture high-level features like edge contours}, as seen in (\textit{d}). }
\label{fig:featureextrac}
\end{figure*}

\begin{figure*} [t!]
	\centering
           
        \subfloat[Input Image]{
	\includegraphics[width=3.3cm,height=2.4cm]{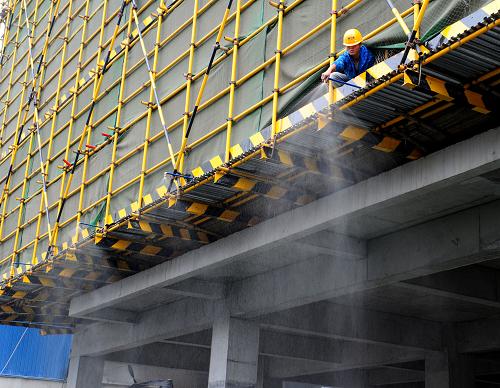}}
        \subfloat[w/o Joint Module]{
	\includegraphics[width=3.3cm,height=2.4cm]{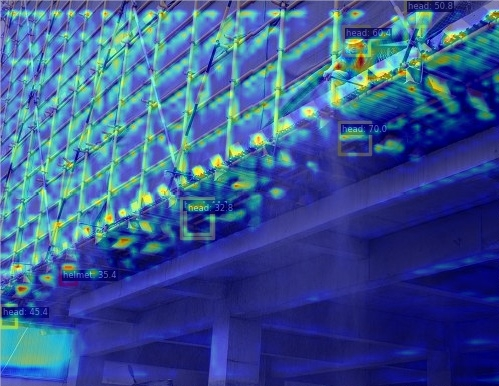}}
         \subfloat[w ECANet]{
	\includegraphics[width=3.3cm,height=2.4cm]{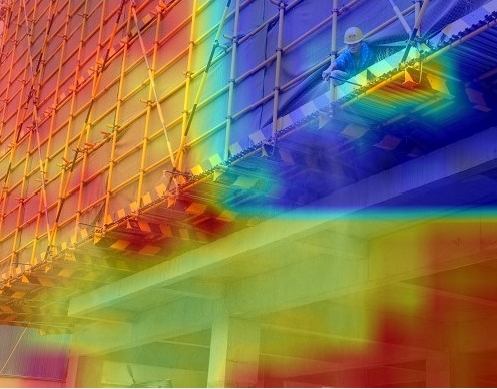}}
          \subfloat[w SC-A]{
\includegraphics[width=3.3cm,height=2.4cm]{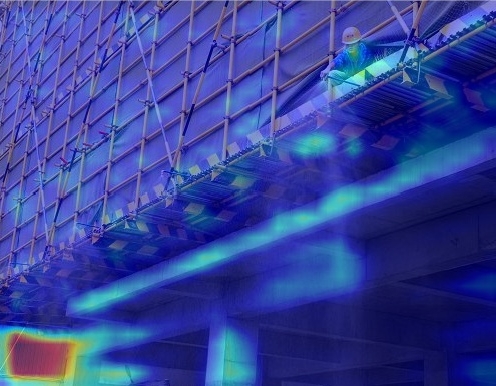}}
           \subfloat[w Joint Module]{
	\includegraphics[width=3.3cm,height=2.4cm]{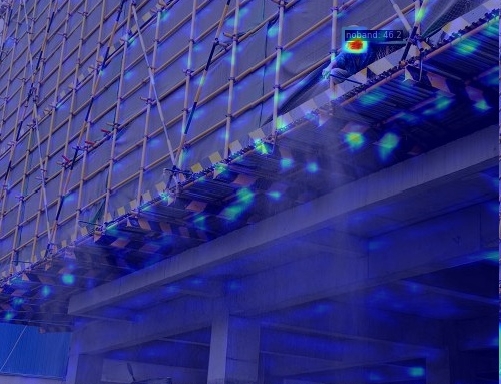}}\\
        \centering{Example I}\\
        
            \subfloat[Input Image]{
	\includegraphics[width=3.3cm,height=2.4cm]{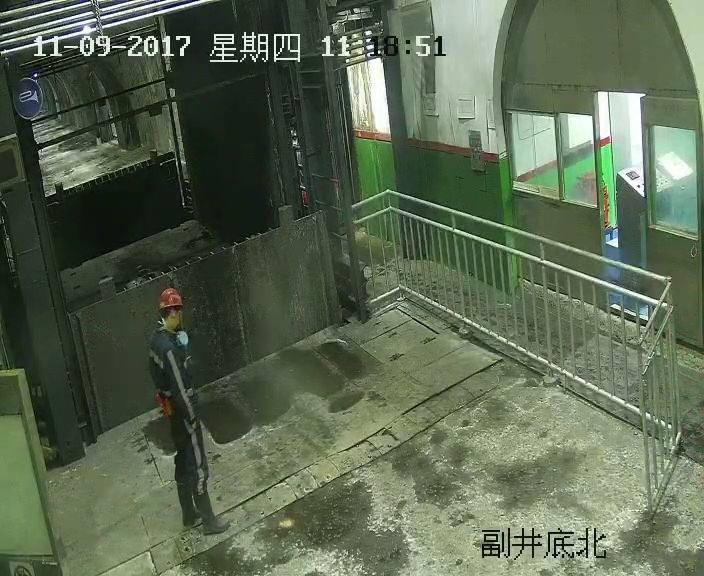}}
         \subfloat[w/o Joint Module]{
		\includegraphics[width=3.3cm,height=2.4cm]{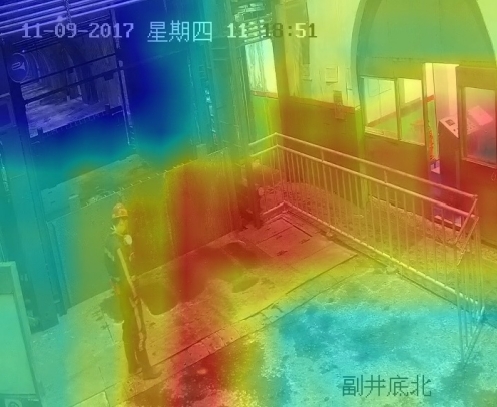}}
         \subfloat[w ECANet]{
		\includegraphics[width=3.3cm,height=2.4cm]{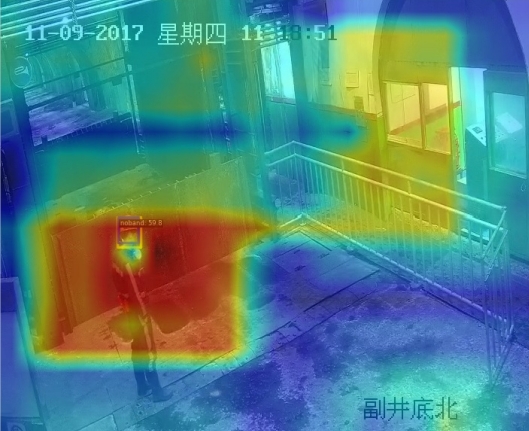}}
         \subfloat[w SC-A]{
		\includegraphics[width=3.3cm,height=2.4cm]{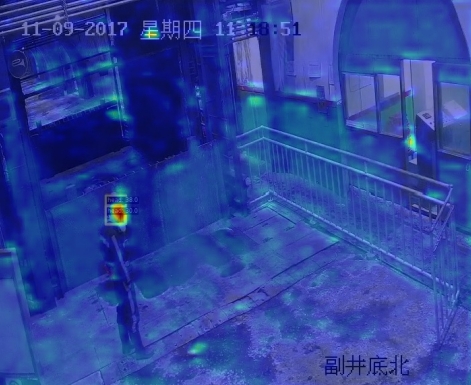}}
         \subfloat[w Joint Module]{
		\includegraphics[width=3.3cm,height=2.4cm]{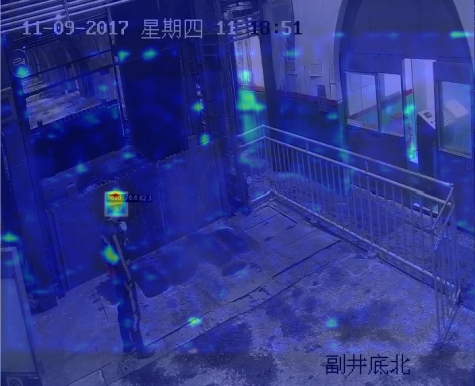}}\\
       
  \centering{Example II}\\
        
  \caption{Comparison of feature perception performance. According to Fig. \ref{fig:YOLODBtoolbox}, when the Neck does not incorporate the Joint Module, the model fails to capture small-sized objects, as shown in (\textit{b}) and (\textit{g}). Even with the Channelwise ECANet, accurate perception cannot be achieved, as shown in (\textit{c}) and (\textit{h}). \textit{With the assistance of SC-A, the model can capture small-sized objects}, as shown in (\textit{i}), but the performance is unstable, as seen in (\textit{d}). \textit{With the full support of the Joint module, the model achieves accurate perception}, as demonstrated in (\textit{e}) and (\textit{j}).}
\label{fig:featurecapt}
\end{figure*}

\textbf{Visualization Results.} As shown in Fig. \ref{fig:coco&hbdd}, with the assistance of ED-TOOLBOX, \textit{ED-YOLO achieves detection performance comparable to that of more complex models, with exceptional accuracy in detecting small and dense objects}. For instance, in image (\textit{g}), ED-YOLO successfully differentiates between ``head" and ``no band", and in image (\textit{u}), it effectively detects the ``person" within a crowd. In contrast, the original YOLOv8-s and two typical lightweight models (YOLOv7-tiny-SiLU and PPYOLOEs) exhibit limited detection capability. Specifically, on the HBDD dataset, they frequently miss small-size targets (e.g., false positives in (\textit{e}) and missed detections in (\textit{f})). On the COCO dataset, they suffer from severe missed detections for dense and occluded targets, as shown in images (\textit{r}), (\textit{s}), and (\textit{t}), where they fail to detect occluded "person" in the crowd, as well as occluded ``bus" and small ``clock" targets. Among the three more complex models (DETR, FCOS-101, and YOLOv7-x), despite some missed detections (e.g., in (\textit{p})), they can nearly detect all small and dense targets.


\textbf{Resource Utilization at Training.} To evaluate the resource utilization of ED-YOLO under the assistance of ED-TOOLBOX during the training phase, we assessed the complexity of the Rep-DConvNet training architecture on the HBDD, as shown in Fig. \ref{fig:rep}, and conducted a GPU memory usage experiment, with the results presented in Fig. \ref{fig:gpu}. Compared to the original YOLOv8-s, ED-YOLO exhibits a significant reduction in GPU memory usage, notably lower than other YOLOv8 versions, with only YOLOv8-n showing slightly lower memory consumption. These results demonstrate that \textit{ED-TOOLBOX effectively reduces model resource consumption during training while maintaining the low complexity of the Rep-DConvNet training structure.}

\subsection{Ablation Study}
\label{sec:Ablation}


We conduct ablation experiments to evaluate the contribution of each component of ED-YOLO to the performance improvement of the backbone YOLO model on the HBDD, as shown in TABLE \ref{tab:abla}. Additionally, we further use visualization results to subjectively assess the contribution of Rep-DConvNet to feature extraction (as shown in Fig. \ref{fig:featureextrac}) and the contribution of SC-A to feature perception (as shown in Fig. \ref{fig:featurecapt}).

\begin{table}[h]
 \setlength{\tabcolsep}{0.1cm}
\centering
    \caption{Results of ablation experiments. When changing any component of our model, the assessment metrics degrade with varying degrees, proving the effectiveness of our improvements.}
\begin{tabular}{cccccc}
\hline
\textbf{Module}                    & \textbf{Algorithm}                     & \textbf{mAP/\%}$\uparrow$ & \textbf{FPS}$\uparrow$    & \textbf{Param/M}$\downarrow$ & \textbf{FLOPs/G}$\downarrow$ \\ \hline
\multirow{1}{*}{Backbone} & w/o Rep-DConvNet  & 85.69  & 90.63  & 13.854  & 28.519  \\
                          \hdashline
\multirow{3}{*}{Neck}     & w/o Joint Module             & 87.31  & 95.68 & 10.925   & 15.802  \\
                          & w/o ECANet          & 90.12  & 94.13 & 10.925  & 19.327  \\
                          & w/o SC-A            & 88.05  & 95.22 & 10.925   & 16.177  \\\hdashline
Head                      & w/o Efficient Head  & 90.68  & 93.15  & 11.241   & 20.885  \\ \hline
\rowcolor{gray!30}&  ED-YOLO                     &  91.34  &  93.84  &  10.925   &  19.725  \\ \hline
\end{tabular}
\label{tab:abla}
\end{table}


\textbf{Numerical Results.} The ablation study results presented in TABLE \ref{tab:abla} demonstrate that \textit{removing any component of ED-TOOLBOX leads to a decline in performance, underscoring the necessity of each module}. Specifically, omitting the Joint module results in a 3.61\% decrease in mAP, highlighting its crucial role in model enhancement. Additionally, excluding SC-A causes a 1.34\% drop in mAP, while removing ECANet leads to a 2.84\% decrease in mAP, illustrating the independent contributions of the spatial and channel attention mechanisms. Replacing Rep-DConvNet with the original backbone significantly reduces mAP by 6.19\% and increases FLOPs by 74.9\%, emphasizing the effectiveness of Rep-DConvNet in balancing accuracy and computational efficiency. Lastly, excluding the Efficient Head results in a 0.72\% reduction in mAP and a 13.2\% increase in FLOPs.


\textbf{Visualization Results.} Fig. \ref{fig:featureextrac} illustrates the output feature maps of the feature extraction network (DarkNet-53) in both shallow and deep layers before and after the improvement with Rep-DConvNet. The results show that \textit{Rep-DConvNet enhances the network’s ability to extract features at different scales}. Specifically, in the shallow layers, compared to the original DarkNet-53, the feature maps extracted by the improved network contain more detailed features, such as the texture of the helmets in the red boxes, as shown in (\textit{b}). This advantage contributes to improved performance in small object detection. Furthermore, in the deep layers, the improved network effectively extracts high-level features such as edges and contours, as shown in (\textit{d}), with superior performance compared to the original network’s results in (\textit{e}).


Additionally, we further evaluate the feature perception capability of the Joint module. As shown in Fig. \ref{fig:featurecapt}, \textit{SC-A enhances the model's performance in small object perception, while the complete Joint module helps the model consistently detect small-sized objects}. Specifically, using ECANet alone does not improve the model’s perception capability, still leading to false detections in (\textit{c}) and (\textit{h}). When the network is configured with SC-A, the perception performance improves significantly, allowing the network to capture the objects in (\textit{i}), although false detections still occur in (\textit{d}). Furthermore, when configured with the complete Joint module, the network can accurately capture the objects compared to the previous networks in (\textit{e}) and (\textit{j}).

\subsection{Visual Surveillance Simulation Experiment.}
\label{sec:Simulation}

In this section, we simulate an IoVT surveillance system follow the experimental setup and deploy ED-YOLO on the edge side to evaluate the practical application value of our ED-TOOLBOX. 

\begin{figure*} [htbp!]
	\centering     
            \subfloat[label image]{
		\includegraphics[width=3.5cm,height=2.6cm]{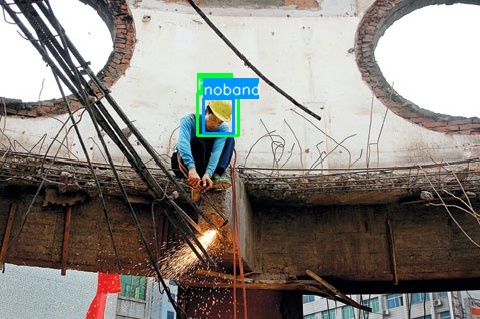}}
         \hspace{-3mm}
             \subfloat[DETR-101]{
		\includegraphics[width=3.5cm,height=2.6cm]{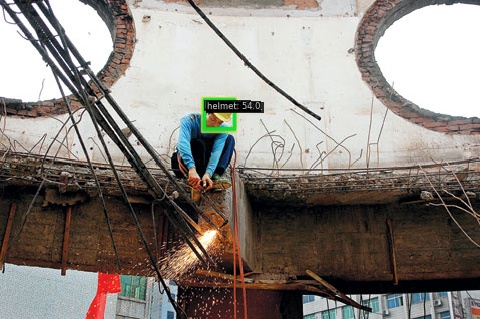}}
         \hspace{-3mm}
         \subfloat[FCOS-101]{
		\includegraphics[width=3.5cm,height=2.6cm]{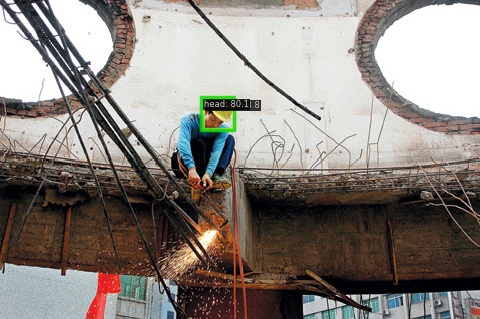}}
         \hspace{-3mm}
            \subfloat[YOLOv7-x]{
		\includegraphics[width=3.5cm,height=2.6cm]{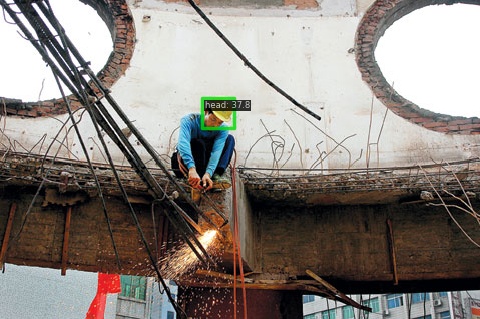}}
  \hspace{-3mm}\\
   \subfloat[YOLOv7-tiny-SiLU]{
		\includegraphics[width=3.5cm,height=2.6cm]{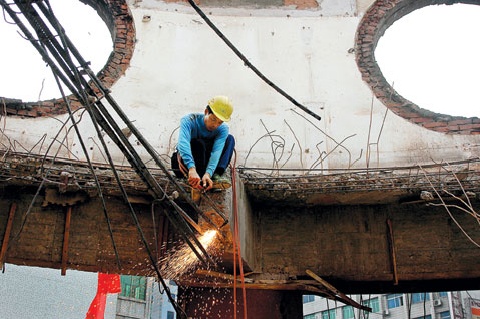}}
         \hspace{-4mm}
             \subfloat[PPYOLOEs]{
		\includegraphics[width=3.5cm,height=2.6cm]{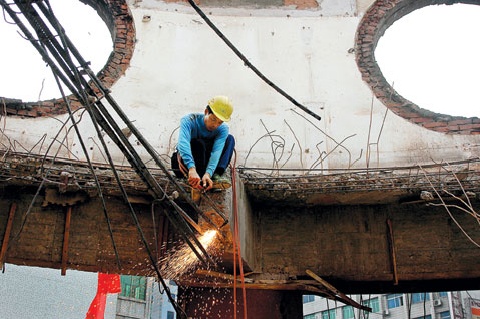}}
         \hspace{-3mm}
         \subfloat[YOLOv8-s]{
		\includegraphics[width=3.5cm,height=2.6cm]{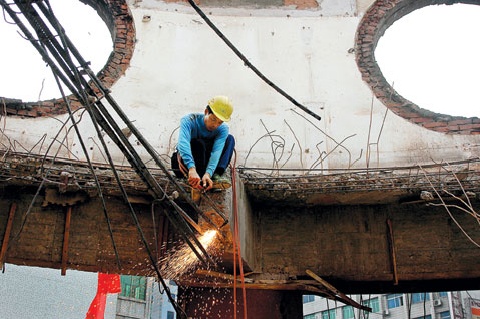}}
         \hspace{-3mm}
            \subfloat[ED-YOLO]{
		\includegraphics[width=3.5cm,height=2.6cm]{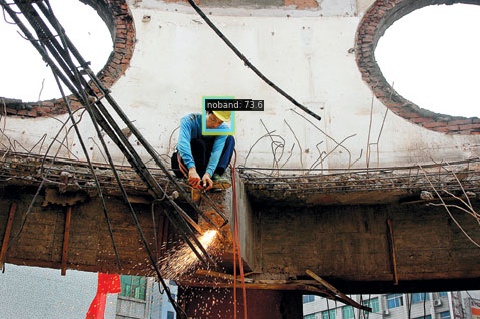}}
  \hspace{-3mm}\\
  \centering{Example I}\\
            \subfloat[label image]{
		\includegraphics[width=3.5cm,height=2.6cm]{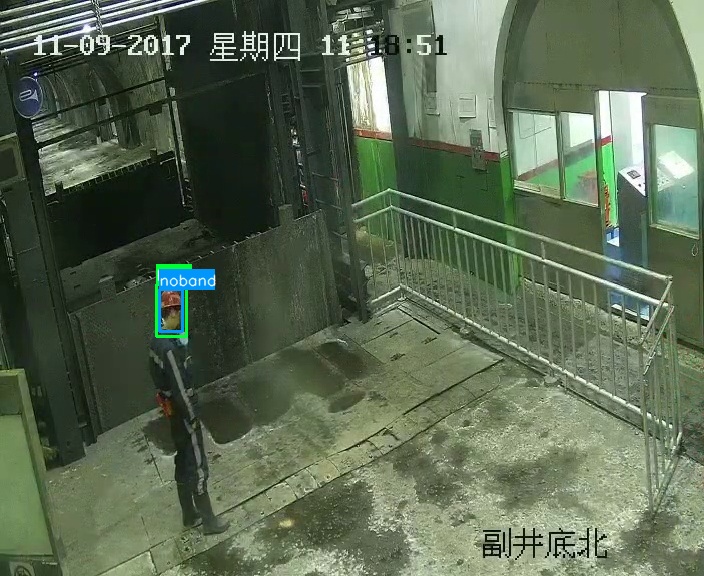}}
         \hspace{-3mm}
             \subfloat[DETR-101]{
		\includegraphics[width=3.5cm,height=2.6cm]{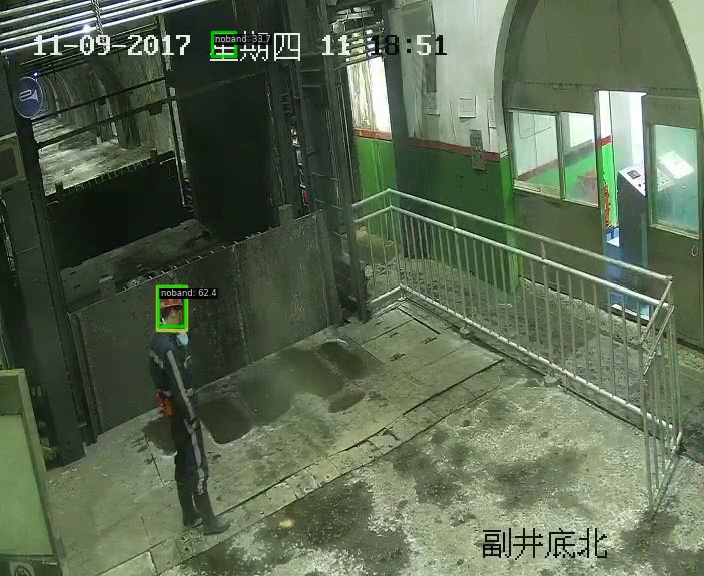}}
         \hspace{-3mm}
        \subfloat[FCOS-101]{
		\includegraphics[width=3.5cm,height=2.6cm]{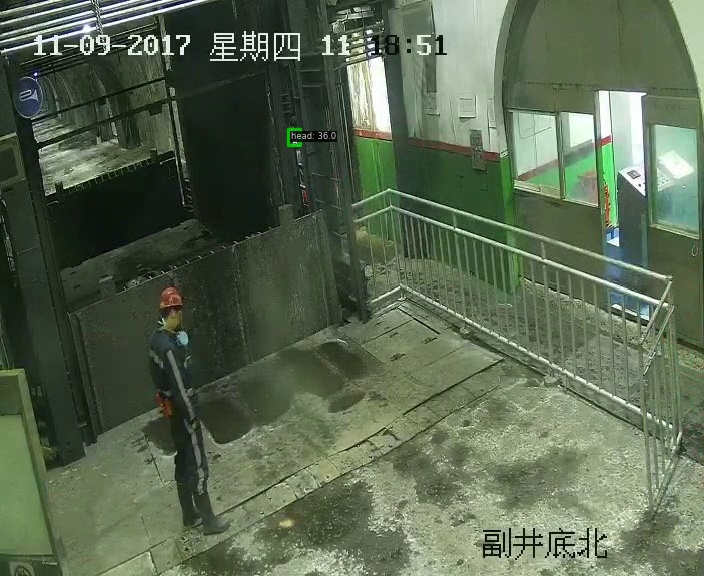}}
         \hspace{-3mm}
            \subfloat[YOLOv7-x]{
		\includegraphics[width=3.5cm,height=2.6cm]{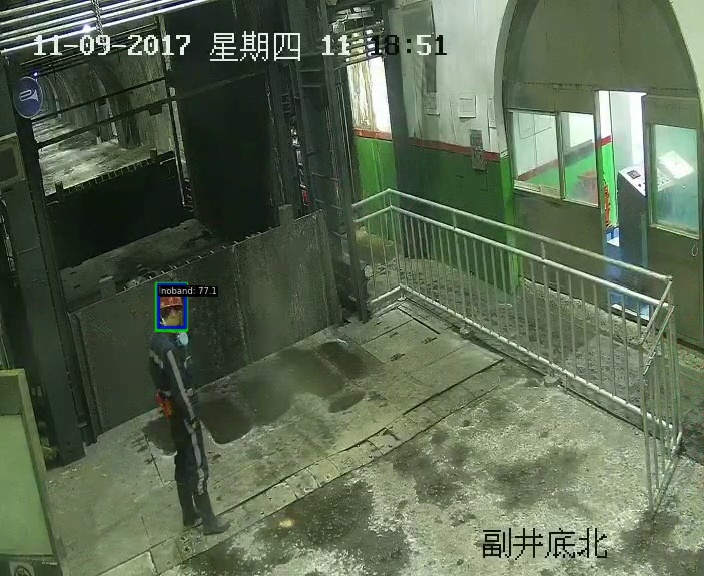}}
  \hspace{-3mm}\\
   \subfloat[YOLOv7-tiny-SiLU]{
		\includegraphics[width=3.5cm,height=2.6cm]{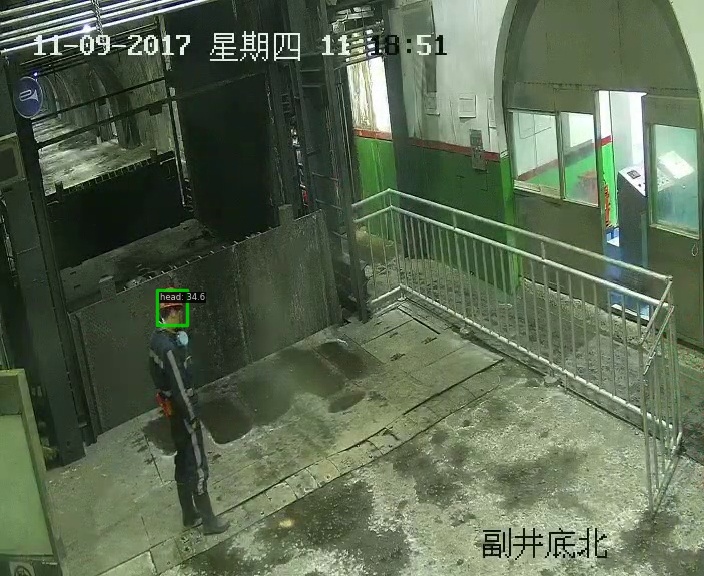}}
         \hspace{-3mm}
             \subfloat[PPYOLOEs]{
		\includegraphics[width=3.5cm,height=2.6cm]{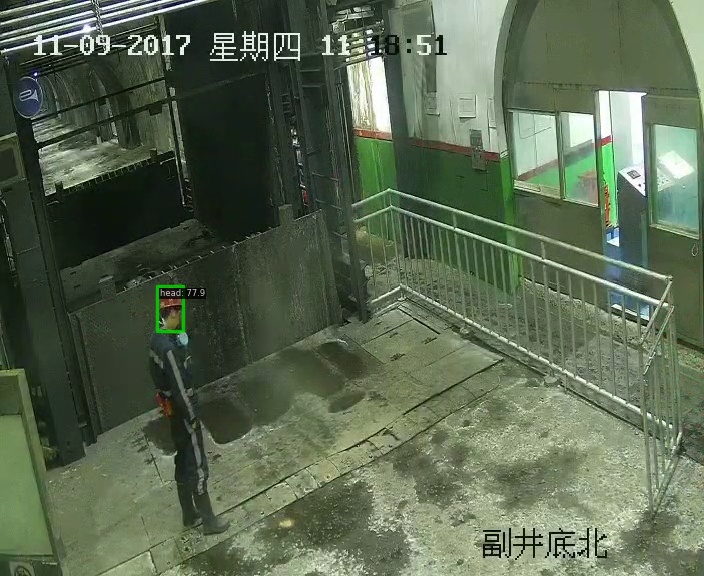}}
         \hspace{-3mm}
         \subfloat[YOLOv8-s]{
		\includegraphics[width=3.5cm,height=2.6cm]{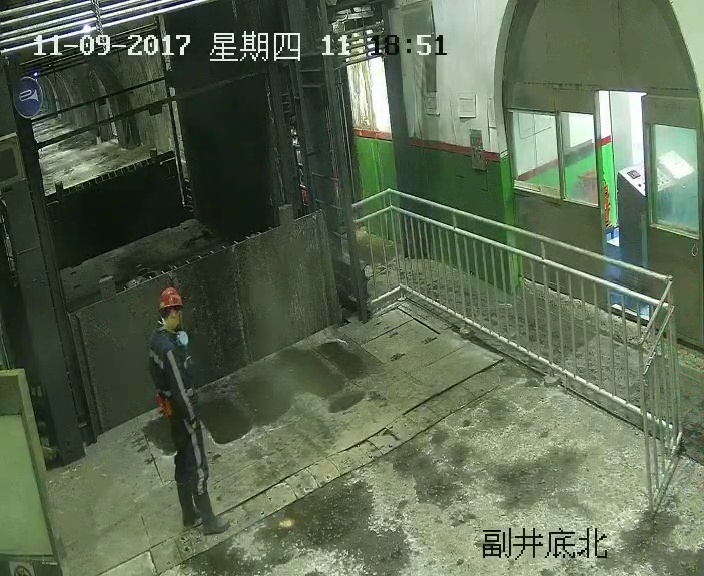}}
         \hspace{-3mm}
            \subfloat[ED-YOLO]{
		\includegraphics[width=3.5cm,height=2.6cm]{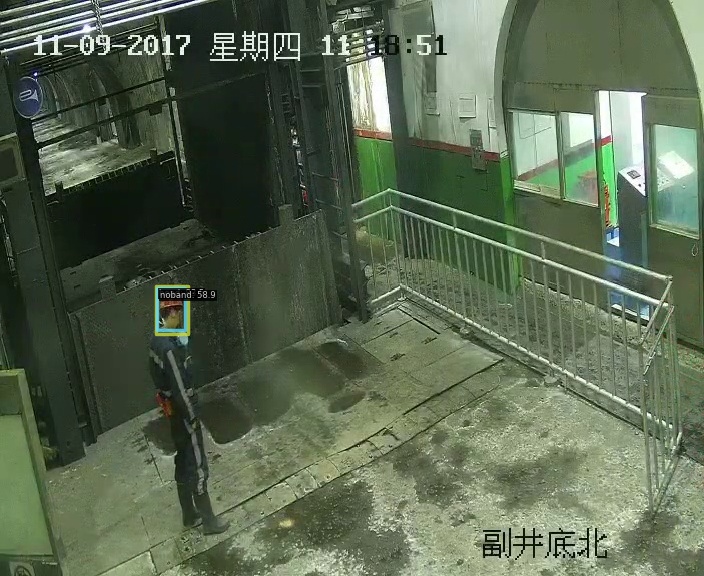}}
  \hspace{-3mm}\\
  \centering{Example II}\\
          
  \caption{Comparison of detection results in the IoVT surveillance system. Compared to other lightweight models, such as YOLOv7-tiny-SiLU, PPYOLOEs, and YOLOv8-s, our \textit{ED-YOLO demonstrates better detection performance}, even surpassing the large model, as shown in (\textit{k}).}
\label{fig:simul}
\end{figure*}



\textbf{Visualization Results.} As shown in Fig. \ref{fig:simul}, in Examples I and II, complex models, such as YOLOv7-x and FCOS-101, result in missed detections (see (\textit{d})) and false detections (see (\textit{k})), indicating that even complex models struggle to capture the fine details of small objects, such as hatbands. The missed detection issue is very common with lightweight models, showing that they are not suitable for small-size hatband detection. In contrast, ED-YOLO is able to accurately detect all targets without missed or false detections, as shown in (\textit{h}) and (\textit{p}), proving that with the assistance of our ED-TOOL, the model has improved capability for small-size hatband detection.

\begin{figure} [b!]
	\centering
\includegraphics[width=0.4\textwidth]{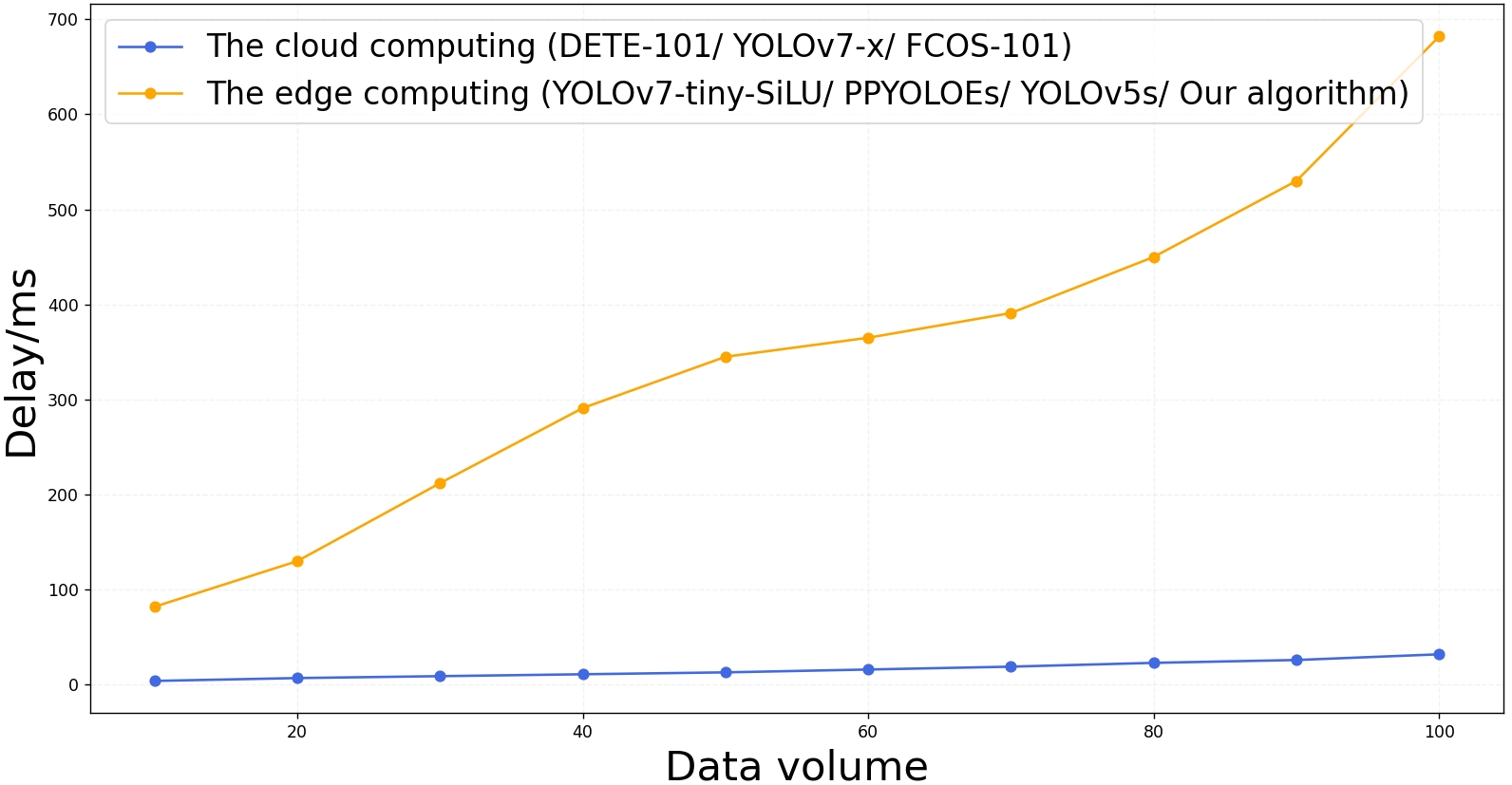}
  \caption{The delay comparison of two computing approaches. The delay difference is huge, and the difference becomes even greater as the amount of transmitted data increases.}
\label{fig:delay}
\end{figure}



\textbf{Numerical Results.} In Fig. \ref{fig:delay}, under our experimental setups, when the number of transmitted images is 50 (approximately the real-time data volume of video surveillance system), the data transmission delay for cloud computing and edge computing is 345ms and 13ms, respectively. Due to the significant delay, the total throughput (TFPS) of cloud computing barely meets the real-time requirements of detection tasks, further illustrating that edge computing is more suitable for time-sensitive tasks. Detailed experimental results are shown in TABLE \ref{tab:simul}.

\begin{table}[h!]
\setlength{\tabcolsep}{0.03cm}
\centering
    \caption{Detection effects of different algorithms. Compared with other methods, our model deployed on the edge side of the system still has satisfactory detection accuracy and real-time inference speed, and is able to provide users with accurate and low-latency helmets detection services. On each dataset, \textcolor{red}{the best results} are marked in red, and \textcolor{blue}{the second-best results} are marked in blue. In the \textbf{lightweight models, the best mAP} is highlighted in bold.}
\begin{tabular}{ccccccc}
\hline
\textbf{Approach}               & \textbf{Method}        & \textbf{mAP/\%}$\uparrow$ & \textbf{FPS}$\uparrow$   & \textbf{TFPS}$\uparrow$  & \textbf{Param/M}$\downarrow$ & \textbf{FLOPs/G}$\downarrow$ \\ \hline

 & DETR-101  \cite{End-to-end}       & 95.47  & 45.4  & 2.73 & 60.1    & 253.8   \\
 
                                 & YOLOv7-x  \cite{YOLOv7}       & \textcolor{red}{96.13}  & 58.6  & 2.76 & 75.6    & 192.3   \\

                               Cloud  & FCOS-101  \cite{FCOS}       & \textcolor{blue}{95.53}  & 70.8  & 2.79 & 52.3    & 132.7   \\\cdashline{2-7}[5pt/1pt]

                               Computing  & YOLOv7-tiny \cite{YOLOv7} & 78.45  & \textcolor{red}{93.3} & 2.81 & \textcolor{blue}{13.8}     & \textcolor{red}{20.5}    \\

                                 & PPYOLOEs  \cite{PP-YOLOE}       & 83.35  & 92.1  & \textcolor{blue}{2.80} & 15.4    & 23.1    \\
 
                                 & YOLOv5s          & 80.19  & \textcolor{blue}{92.7} & \textcolor{red}{2.81} & \textcolor{red}{13.1}     & \textcolor{blue}{21.5} \\\hline\hline

 & DETR-101  \cite{End-to-end}       & 93.86  &   -   & -      & 60.1    & 253.8   \\

                                 & YOLOv7-x \cite{YOLOv7}        & \textcolor{red}{94.19}  & 2.5     & 2.42       & 75.6    & 192.3   \\
 
                             Edge     & FCOS-101  \cite{FCOS}       & \textcolor{blue}{94.08}  & 6.6    & 6.08       & 52.3    & 132.7   \\\cdashline{2-7}[5pt/1pt]

                              Computing   & YOLOv7-tiny \cite{YOLOv7}& 76.53  & \textcolor{blue}{21.2}     & \textcolor{blue}{16.64}       & 13.8     & \textcolor{blue}{20.5}    \\

                                 & PPYOLOEs   \cite{PP-YOLOE}      & 82.17 & 20.1     & 15.94       & 15.4     & 23.1    \\

                                 & YOLOv5s          & 79.55  & 20.7     & 16.31       & \textcolor{blue}{13.1}     & 21.5  \\   
                            \rowcolor{gray!30} ED-YOLO        & \textbf{90.21}  & \textcolor{red}{21.6}     & \textcolor{red}{16.86}      & \textcolor{red}{10.9}     & \textcolor{red}{19.7}    \\ \hline
\end{tabular}
\label{tab:simul}
\end{table}

In TABLE \ref{tab:simul}, in the cloud computing setup, all models are impacted by system latency (as shown in Fig. \ref{fig:delay}), with the total throughput (TFPS) ranging from 2.71 to 2.83, which is far below the real-time detection requirements. This further demonstrates that cloud computing is insufficient to meet the demands of practical detection tasks. In the edge computing setup, although larger models experience a slight decrease in accuracy due to computational limitations, they still maintain superior performance. However, resource-constrained embedded devices are unable to support large models for real-time inference, with some models, such as DETR-101, failing to work altogether. Lightweight models like YOLOv7-tiny, PPYOLOEs, and YOLOv5s can achieve relatively high inference speeds on edge devices, but their low detection accuracy does not meet the required standards. In contrast, \textit{our ED-YOLO can be deployed on embedded devices for real-time inference, with detection accuracy similar to that of large models}. These results show that the detection model assisted by ED-TOOL can perform accurate and real-time hatband detection on IoVT edge devices, proving that \textit{in real-world tasks, ED-TOOL enables edge deployment of detection models while maintaining outstanding performance, highlighting the practical application value of our ED-TOOL.}




\section{Conclusion \& Future Works} 
\label{sec:conlusion}


This study presents \textit{the Edge Detection Toolbox (ED-TOOLBOX), specifically designed for edge deployment of detection models, featuring plug-and-play components that efficiently deploy detection models while maintaining their performance}. We introduce the \textit{Reparameterized Dynamic Convolutional Network} (Rep-DConvNet), specifically designed to maintain the feature extraction of detection models, improving detection performance while reducing model size. Additionally, to address the lack of connectivity between modules, we propose a \textit{Joint Module} based on \textit{Sparse Cross-Attention} (SC-A), enabling adaptive key features transfer with low computational complexity. For YOLO-based models, we design an \textit{Efficient Head} to optimize edge deployment performance over the original decoupled head. Furthermore, recognizing that current helmet detection tasks overlook the importance of band fastening, we \textit{introduce the Helmet Band Detection Dataset (HBDD) and apply the edge detection models to address this practical issue}. Through extensive experiments, we have validated the effectiveness and practical value of ED-TOOLBOX. However, the following limitations remain:

\begin{itemize} \item\textbf{Incompatibility with Transformer-based detection models}. Current components are designed for CNN-based models \cite{ali2024yolo, liu2016ssd}, but are not compatible with Transformer-based detection models \cite{li2023transformer}.

\item\textbf{Limited application domain}. ED-TOOLBOX currently does not support other tasks, such as instance segmentation \cite{xie2022edge} and object tracking \cite{zhao2020iot}, which also require edge deployment.
\end{itemize}

Future work will focus on addressing these issues. \textbf{First}, we plan to leverage the sparse cross-attention (SC-A) structure and our multi-scale feature attention expertise \cite{wu2024small} to \textit{develop lightweight, plug-and-play local feature mapping components optimized for Transformer architectures}. \textbf{Second}, we plan to \textit{develop dedicated edge deployment components and general modules applicable to multiple visual models}. Importantly, \textit{similar to foundational models \cite{bommasani2021opportunities,zhang2024challenges}}, users can fine-tune the pre-trained general module to adapt it to specific tasks. In summary, we aim to expand the “\textit{Edge TOOLBOX}” ecosystem to meet the growing requirements of edge computing.






\bibliographystyle{cas-model2-names}
\bibliography{ref.bib}

\end{document}